%% file: iclr2026_conference.tex
\documentclass{article} % For LaTeX2e
\usepackage{iclr2027_conference,times}

\input{math_commands.tex}

\usepackage{hyperref}
\usepackage{url}
\usepackage{graphicx}
\usepackage{subcaption} % for subfigure environments
\usepackage{placeins} % provides \FloatBarrier
\usepackage{amssymb}
\usepackage{algorithm}
\usepackage{algpseudocode}
\usepackage[inkscapelatex=false]{svg}

\title{The Evolution Strategies Paradox:\\Better Models from Worse Perturbations}

\author{%
\parbox{\dimexpr\textwidth-2\tabcolsep\relax}{\raggedright
Roberto Dailey$^{1}$\thanks{Corresponding author: \texttt{roberto.dailey@cognizant.com}},
Kajetan Schweighofer$^{1}$, Jamieson Warner$^{1}$, Conor F.~Hayes$^{1}$,
Elliot Meyerson$^{1}$, Babak Hodjat$^{1}$, Risto Miikkulainen$^{1,2}$, Xin Qiu$^{1}$} \\
$^{1}$Cognizant AI Lab, San Francisco \quad $^{2}$The University of Texas at Austin, Austin
}

\iclrfinalcopy  %Uncomment for camera-ready version, but NOT for submission.
\begin{document}

\maketitle

\begin{abstract}
How is it possible for perturbation-based methods like Evolution Strategies (ES) to optimize billion-parameter large language models (LLMs)? Classical search methods are highly dependent on finding good candidates within generated populations. However, when using ES to fine-tune LLMs, a phenomenon can occur where all offspring models consistently perform worse than the parent model, yet during training, the parent model consistently improves. This paradox occurs across models, tasks, and even modalities. This paper characterizes the paradox, then examines how the local reward landscape and properties of the ES estimator contribute to it: in high-dimensional spaces, exploration leads to worse performance, yet still highlights good directions; aggregating those directions then allows for making local improvements.  The result can be seen as a new optimization paradigm where a principled interaction of noise and curvature produces the paradox as a byproduct of effective optimization.  These findings provide guidance for taking advantage of perturbation-based methods in LLM fine-tuning and in high-dimensional optimization in general.
\end{abstract}

\section{Introduction}
Optimization using population-based search, i.e.\ evolutionary computation, relies on generating and selecting high-quality candidates. Genetic algorithms optimize by discovering beneficial recombinations and mutations along with selection of candidates with higher fitness \citep{ga}. Particle swarm optimization pressures particles to search for better positions while maintaining the previously best found positions \citep{particle_swarm}. The original evolution strategies (ES) compared a single parent to a single offspring and selected the better solution \citep{rechenberg1965cybernetic}. In each of these methods, the generation of higher-fitness candidate solutions is crucial to the optimization process.

For ES specifically, this story has changed over time. With the introduction of natural evolution strategies \citep[NES,][]{natural_es}, which updates the parent model using a weighted sum of all generated candidates, ES was shown to potentially behave as a gradient approximator instead of working through the selection of candidates with an improved fitness. NES was a precursor to implementation of ES on large-scale reinforcement learning (RL) problems such as Atari \citep{es_openai} that then inspired theoretical analysis that found that ES with isotropic Gaussian noise is a gradient estimate: with high enough population sizes (ranging from thousands to hundreds of thousands), correlation between the ES update direction and network gradient stays consistent while ES performance on a per-iteration basis would match that of backpropagation \citep{mnist_evolution_strategies}. This along with further work concluded that ES operated as a gradient estimator that naturally favors robustness \citep{lehman2018more}.

\begin{figure}[t]
    \centering
    \includegraphics[width=0.95\linewidth]{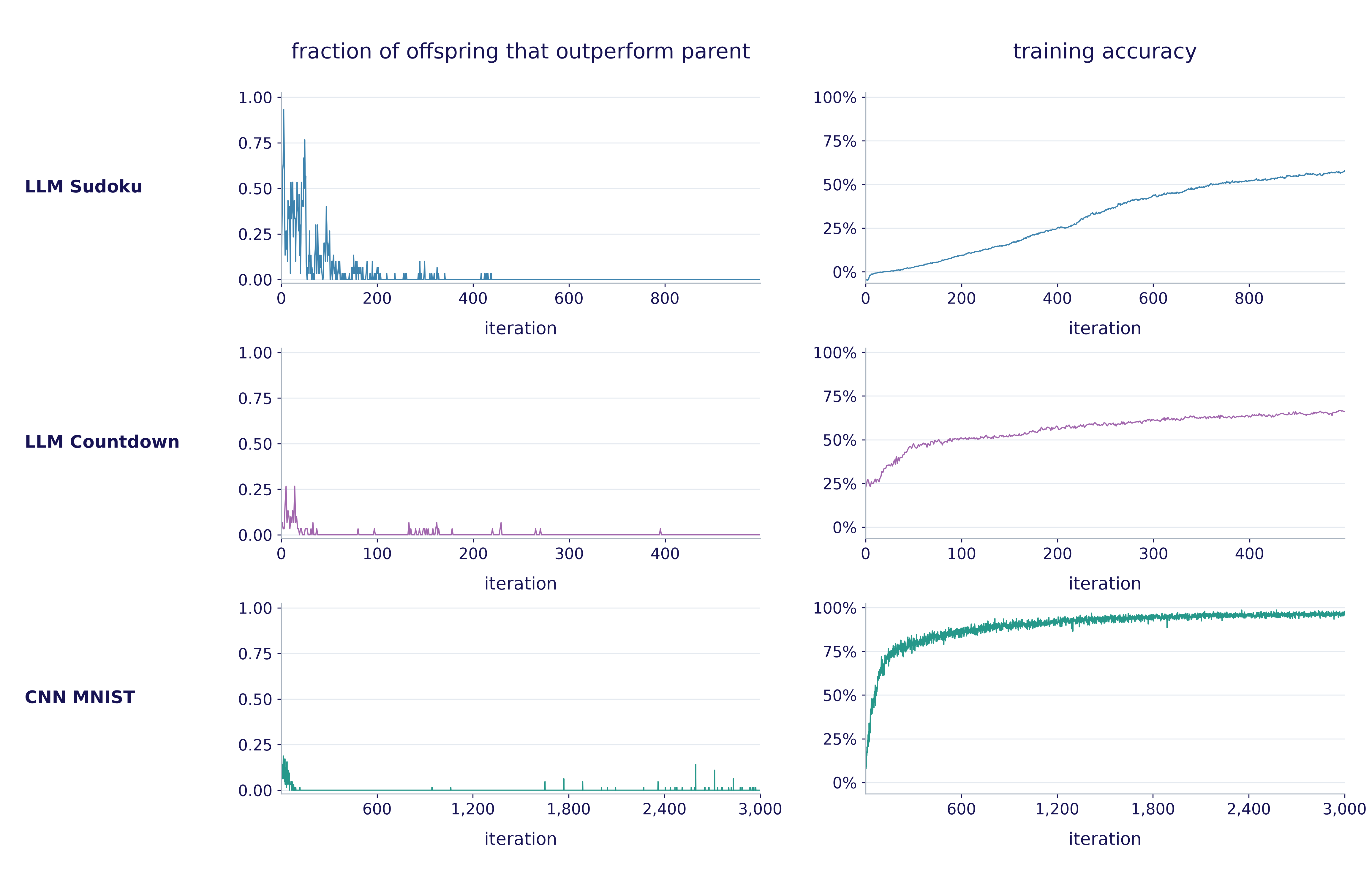}
    \caption{The Evolution Strategies Paradox across three examples: 
    OLMo 2 7B fine-tuned on Sudoku, Qwen 2.5 7B fine-tuned on Countdown, and a 60K parameter CNN on MNIST. Left plots show the fraction of offspring models better than the parent model; right plots show the training curves (i.e.\ the performance of the parent models over time). In each case, offspring models perform worse than the parent model for nearly all of training, while the parent models consistently improve.}
    \label{fig:paradox_examples}
\end{figure}

Recently, a new opportunity for ES has emerged: scaling to the vastly larger optimization problem of fine-tuning Large Language Models \citep[LLMs;][]{qiu2026evolutionstrategiesscalellm}. This work shifted from centered rank rewards to z-score normalized rewards, but more surprisingly, it brought the population size down from thousands to just 30 \citep{qiu2026evolutionstrategiesscalellm,hayes2026bestguessimprovingllm}. With such small populations in a space of billions of parameters, accurate gradient estimation appears infeasible.
It was further demonstrated that large pretrained models have a rich search landscape with many directions for improvement \citep{liang2026blessingdimensionalityllmfinetuning,gan2026neural}. 
\begin{figure}[t]
    \centering
    \begin{subfigure}[t]{0.40\linewidth}
        \centering
        \includegraphics[width=\linewidth]{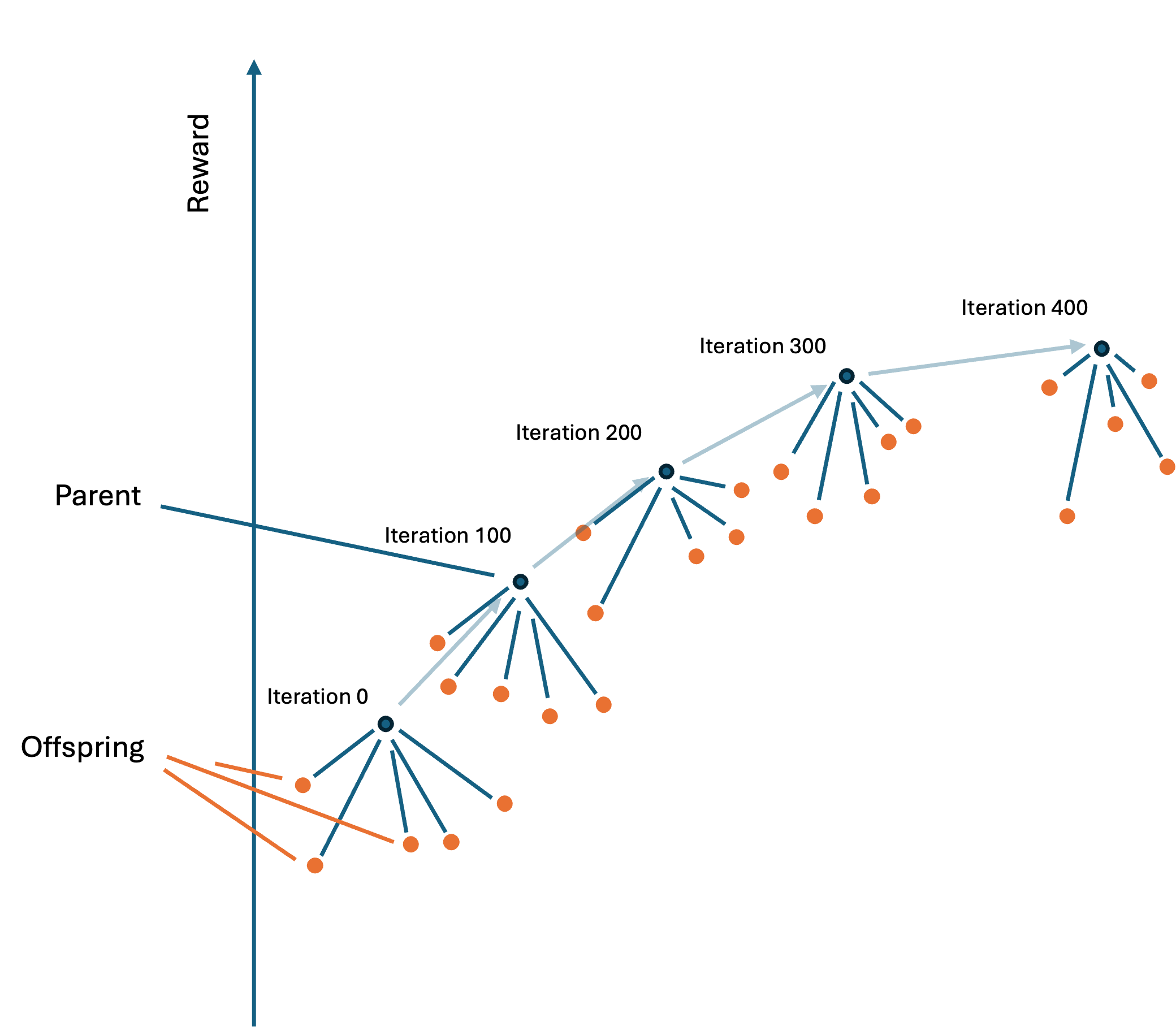}
        \caption{Visual representation of ES Paradox}
        \label{fig:paradox_tracking_visual}
    \end{subfigure}\hfill
    \begin{subfigure}[t]{0.40\linewidth}
        \centering
        \includegraphics[width=\linewidth]{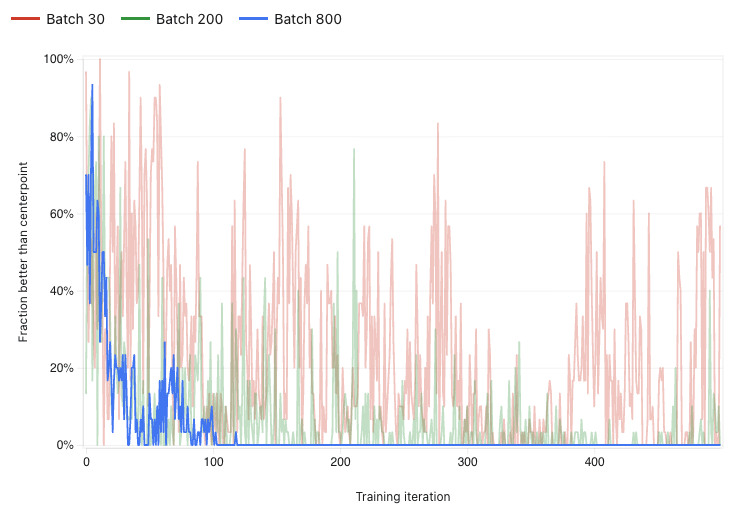}
        \caption{Fraction of population members outperforming parent at various batch sizes.}
        \label{fig:paradox_tracking_batch}
    \end{subfigure}
    \caption{Tracking the ES paradox: (a) A visual representation of the ES paradox over iterations in a training run. At each iteration, offspring are generated which are all worse than the parent model in measured reward. (b) Fraction of perturbations better than the parent in the Sudoku fine-tuning task over iterations in the training run. Larger batch size leads to less noise in the reward landscape, and makes the paradox more prominent.}
    \label{fig:paradox_tracking}
\end{figure}

Taken together, these results suggest that in LLM-scale spaces, ES is not based on gradient estimation, but relies on finding good candidates among sampled perturbations and moving towards them, as other evolutionary methods do.

However, the results in this paper contradict that hypothesis as well: ES is found to make consistent progress even when all offspring candidates are worse than the parent (Figure~\ref{fig:paradox_examples}). This paradox is observed across a range of ES optimization tasks, suggesting that it is a general property of high-dimensional ES optimization. While it had been hypothesized possible for individual steps with alternative forms of ES under idealized conditions \citep{ba2026understandingevolution}, this work is the first empirical demonstration of this occurring. A further analysis reveals that this phenomenon results from three mechanisms 
(1) Consistent curvature of the search landscape results in decreased performance with increased perturbation size; (2) relative performance between offspring still differentiates good and bad directions in parameter space; and (3) the ES preserves the signal of good directions while reducing noise in the aggregated update (Figure~\ref{fig:paradox_tracking}\subref{fig:paradox_tracking_visual}).

The remainder of the paper is as follows: Section~\ref{sec:es_paradox} details the ES update and the experimental setup in LLM fine-tuning under which the paradox is observed. The next three sections then focus on each of the three mechanisms. Section~\ref{sec:landscape} runs a landscape analysis, finding that increased perturbation size is strongly tied to a decrease in reward. Section~\ref{sec:correlation} demonstrates that even at larger perturbations where all offspring are worse than the parent, good directions can still be identified. Section~\ref{sec:es_update} uses an analytical model of ES to show that the updates aggregate those good directions while reducing noise, making the paradox likely. These findings thus demonstrate a previously unknown mechanism, indeed a paradigm shift in optimization, that allows scaling perturbation-based optimization to modern LLMs.

%New section order

\section{The Evolution Strategies Paradox}
\label{sec:es_paradox}
%\subsection{preliminaries}

This work uses ES as implemented by \citet{qiu2026evolutionstrategiesscalellm}, which is characterized by isotropic Gaussian perturbations, low population size, and z-score normalized rewards. For each ES iteration $t$, let a \emph{parent model} be parameterized by current parameters $\theta_t \in \mathbb{R}^d$. A set of $N$ perturbation vectors, indexed $i \in \{1,2,...N\}$ is drawn from isotropic gaussian noise, $\epsilon_i \sim \mathcal{N}(0, I_d)$. Each perturbation is then multiplied by perturbation scale $\sigma \in \mathbb{R}^+$, and added to $\theta_t$ to create a population of \emph{offspring models}, parameterized by $\theta_t + \sigma \epsilon_i$. Each offspring model is evaluated to obtain a reward $R_i$, and the ES update is then the average of the perturbations, weighted by the z-score of the model rewards:

\begin{equation}
    \Delta\theta \;=\; \theta_{t+1} - \theta_t \;=\; \frac{\alpha}{N}\sum_{i=1}^N Z_i\, \epsilon_i ,
    \qquad
    Z_i \;=\; \frac{R_i - \mu_R}{\sigma_R},
    \label{eq:es-update}
\end{equation}

where $\alpha$ is the step size, $\mu_R$ and $\sigma_R$ are the mean and standard deviation of this population's rewards, and weights $Z_i$ are the z-score normalized rewards. These updates are iterated until a stopping criterion, as shown in the basic ES training Algorithm~\ref{alg:es-llm_general}. For an iteration $t$, let $A_t$ denote the event that all offspring at iteration $t$ perform worse
than their parent. The ES paradox occurs when the update still improves the parent model in expectation under this event:
\begin{equation}
  A_t = \Big\{ R\big(\theta_t + \sigma \epsilon_i\big) < R(\theta_t) \;\; \forall\, i \in \{1, \dots, N\} \Big\},
  \qquad
  \mathbb{E}\big[\, R(\theta_{t+1}) - R(\theta_t) \,\big|\, A_t \,\big] > 0.
  \label{eq:es-paradox}
\end{equation}

In training runs, the paradox is tracked by recording the fraction of perturbations with rewards better than the current parent on the training reward. When this metric stays at or close to zero for extended periods of time while reward increases, it is assumed the paradox is occurring. In this paper, the performance on training data is treated as the full reward landscape; however, the effect can still occur when measured with a held-out reward. Monitoring this parent reward adds a marginal inference cost to ES training runs, as one additional model evaluation must be run for each sampled population. Figure~\ref{fig:paradox_examples} displays this metric alongside training progress in fine-tuning in the three example tasks.
Each training run starts with significant fractions of perturbations outperforming the parent model, but partway into training, this metric falls to zero. The effect is most prominent with a fixed batch, though it can occur in shuffled batches; all else equal larger batch sizes make the effect stronger, as shown in Figure~\ref{fig:paradox_tracking}\subref{fig:paradox_tracking_batch}.

This paper's analysis focuses on LLM fine-tuning tasks, where it was first discovered. Two examples of the paradox in LLM fine-tuning are more carefully analyzed: a 500-iteration training run on a $4\times 4$ Sudoku task, and a 500-iteration training run on a countdown task \citep{tinyzero}, using harnesses and grading used in \citet{dtree}. In both these examples, Qwen 2.5 7B-Instruct was used as the LLM, evaluated at temperature zero; ES was run with hyperparameters (N = 30, $\sigma$ = 0.001, $\alpha$ = 5 × 10$^{-4}$). Fine-tuning was done with a dataset of 800 problems, with the batch size set to the entire dataset to demonstrate a strong effect (Figure~\ref{fig:paradox_tracking}\subref{fig:paradox_tracking_batch}). Although the analysis focused on Qwen 2.5 with these two specific tasks, the paradox has been observed in other models as well, such as OLMo 2 7B, as shown in Figure ~\ref{fig:paradox_examples}.
%\ref{fig:paradox_examples}. 

In addition to LLM fine-tuning, the ES paradox occurs in other domains of ES optimization of neural networks. One such domain is included in this paper: training of a randomly initialized 60,000-parameter Convolutional Neural Network (CNN) on MNIST. The paradox was reproduced with ES parameters N = 16, $\sigma$ = 0.04, and $\alpha$ = 0.02. Thus, the paradox can be seen as a general phenomenon across models, tasks, and modalities. The next three sections analyze the mechanisms that produce it and demonstrate that it is a side effect of effective optimization in high-dimensional search spaces.

%This paper will focus on examples of the paradox within LLM fine-tuning tasks, but will demonstrate empirical observations in additional machine learning problems such as MNIST.

\begin{algorithm}[h]
	\caption{Basic ES Algorithm}
	\label{alg:es-llm_general}
	\begin{algorithmic}[1]
		\Require Pretrained LLM with initial parameters $\theta_0$, reward function $R(\cdot)$, total iterations $T$, population size $N$, noise scale $\sigma$, learning rate $\alpha$.
		\For{$t = 1$ to $T$} \Comment{outer ES iterations over generations of populations}
		\For{$i = 1$ to $N$} \Comment{inner ES iteration over population members}
		\State Sample noise $\bm{\varepsilon}_i \sim \mathcal{N}(0,I_d)$
		\State Compute reward for perturbed parameters: \Statex\hspace{\algorithmicindent}\hspace{\algorithmicindent}$R_i\leftarrow R(\theta_{t-1}+\sigma \cdot \bm{\varepsilon}_i)$
		\EndFor
        \State Z-score normalization of rewards
		\Statex\hspace{\algorithmicindent}$Z_i \leftarrow \frac{R_i - \mu_R}{\sigma_R}$
		\State Update model parameters as 
        \Statex\hspace{\algorithmicindent}$\bm{\theta}_{t} \leftarrow \bm{\theta}_{t-1}+ \alpha \cdot \frac{1}{N}\sum_{i=1}^N Z_i \bm{\varepsilon}_i$
		\EndFor
	\end{algorithmic}
\end{algorithm}

%\section{Reward Landscape Analysis via Magnitude Distribution Tests}
\section{Why Are Offspring Worse Than the Parent?}
\label{sec:landscape}

\begin{figure}
    \centering
    \includegraphics[width=1.0\textwidth]{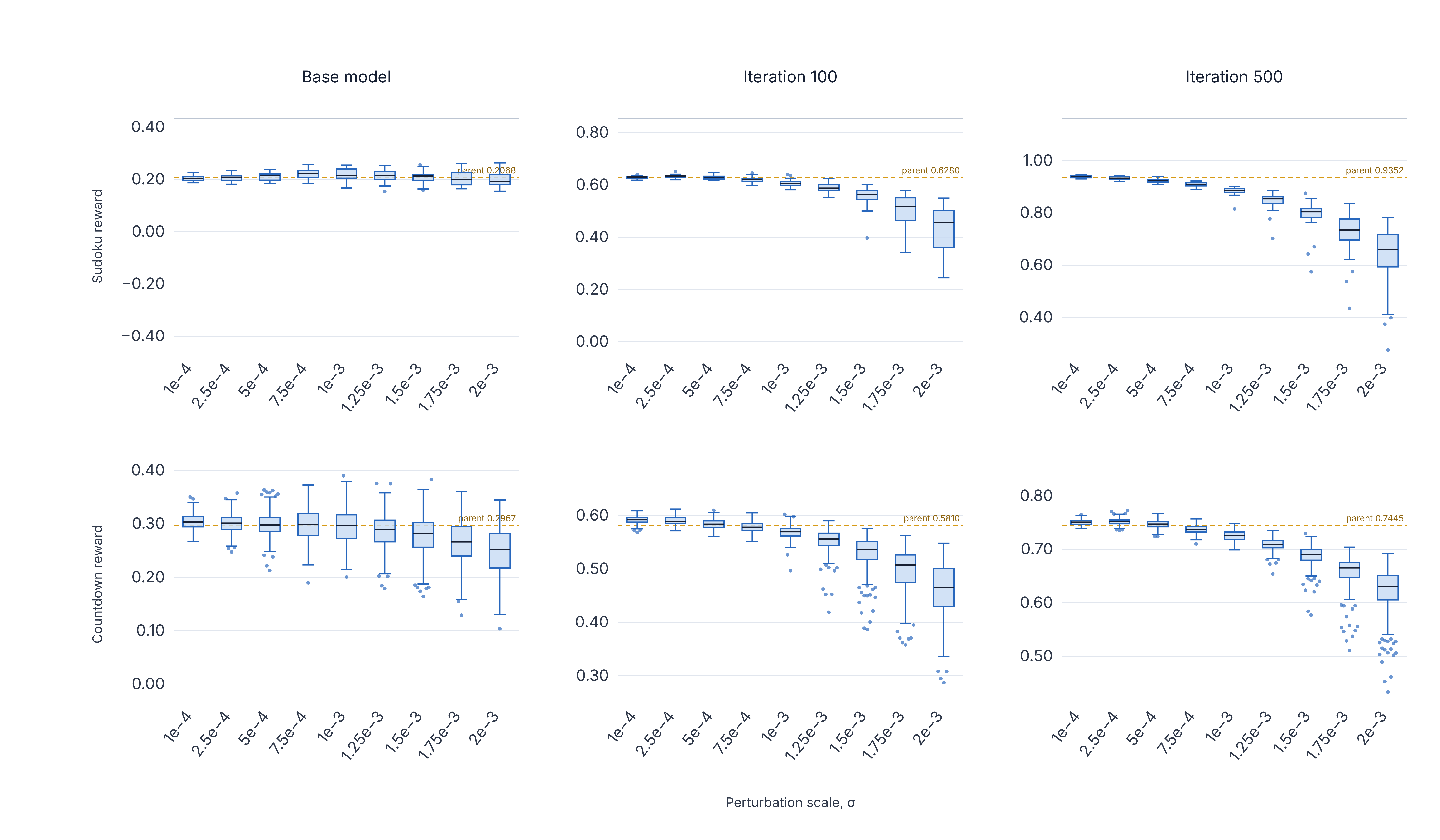}
    \caption{Reward distributions at different perturbation magnitudes. Checkpoints in Countdown and Sudoku training runs were perturbed 300 times at each specified magnitude, and distributions of rewards were plotted. As the magnitudes increase, the means decrease, and the variance increases. Thus, the offspring are likely to perform worse than the parent.}
    \label{fig:magnitude_tests}
\end{figure}

As a first step to understand how the ES paradox emerges, this section examines the relationship between perturbation size and reward in fine-tuning tasks. For both Sudoku and Countdown fine-tuning tasks, checkpoints from every 100 iterations were taken, and 300 offspring were generated at varying perturbation magnitudes. These perturbed models were then evaluated on the original fixed training set of 800 problems, producing a distribution of model rewards at each magnitude. The magnitudes were chosen to match the average magnitude of perturbations with $\sigma = 10^{-4}$ and $\sigma$ from $2.5 \times 10^{-4}$ to $2 \times 10^{-3}$ in steps of $2.5 \times 10^{-4}$. Figure~\ref{fig:magnitude_tests} displays these tests at the base model and at iterations 100 and 500 in the training run, with box-and-whisker plots displayed at each perturbation magnitude. 

With the base model, the mean reward is largely constant, but its variation increases approximately linearly as perturbation size increases. By iteration 100, perturbation magnitude has a significant, consistent, negative effect on the reward. By iteration 500, this effect is even more consistent and pronounced. In Sudoku in particular, the rewards at magnitude matching the $\sigma$ used for training (1e-3) are multiple standard deviations below those of the parent model. Therefore, the perturbation reward distribution strongly depends on the perturbation size ($\sigma$). Small $\sigma$'s produce valid offspring for model improvement, whereas large $\sigma$ nearly guarantees that no better models will be produced. 

\begin{figure}
    \centering
    \begin{subfigure}[t]{0.40\linewidth}
        \centering
        \includegraphics[width=\linewidth]{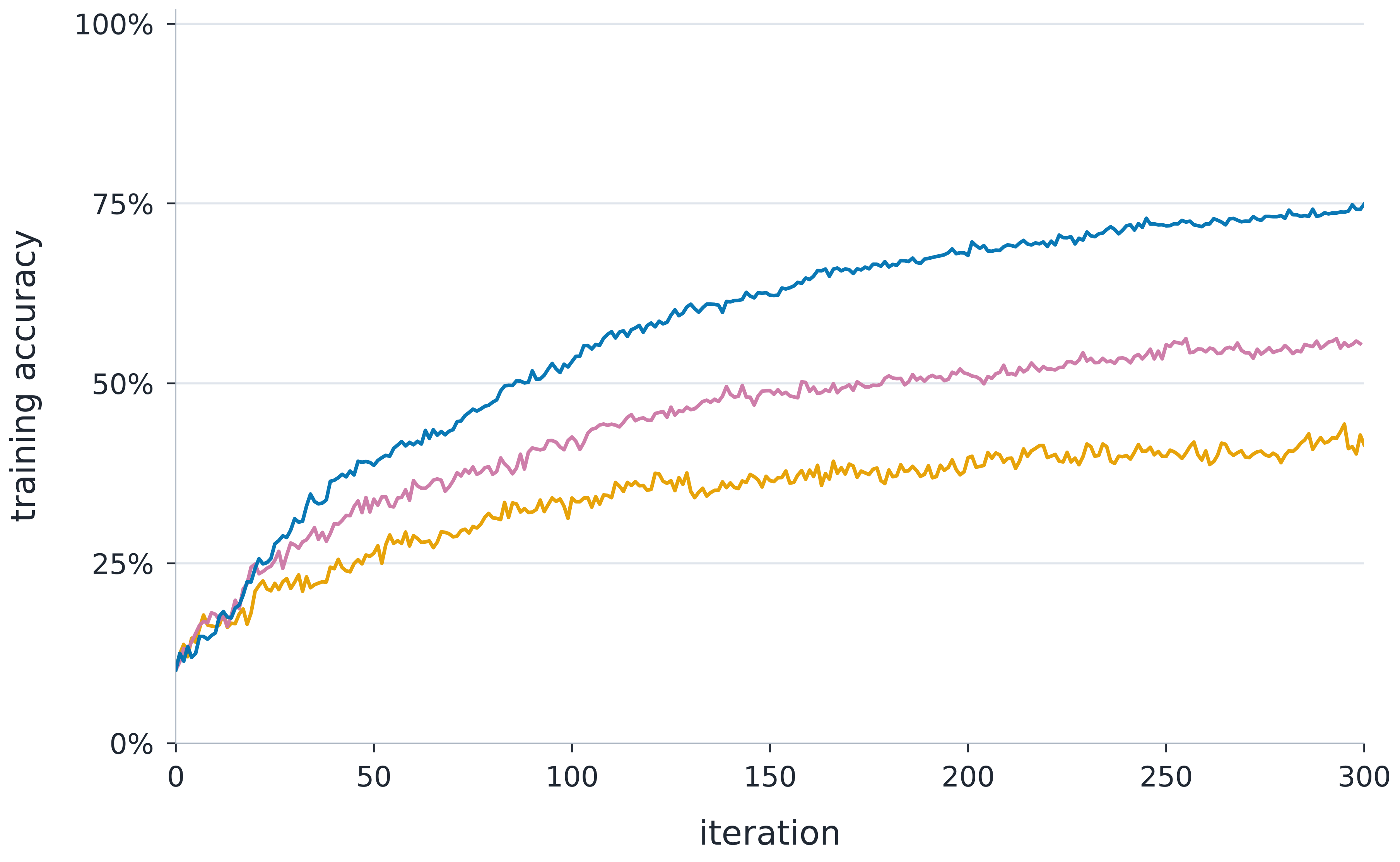}
        \caption{Fine-tuning performance on Sudoku}
        \label{fig:small_sigma_comparison_train}
    \end{subfigure}\hfill
    \begin{subfigure}[t]{0.40\linewidth}
        \centering
        \includegraphics[width=\linewidth]{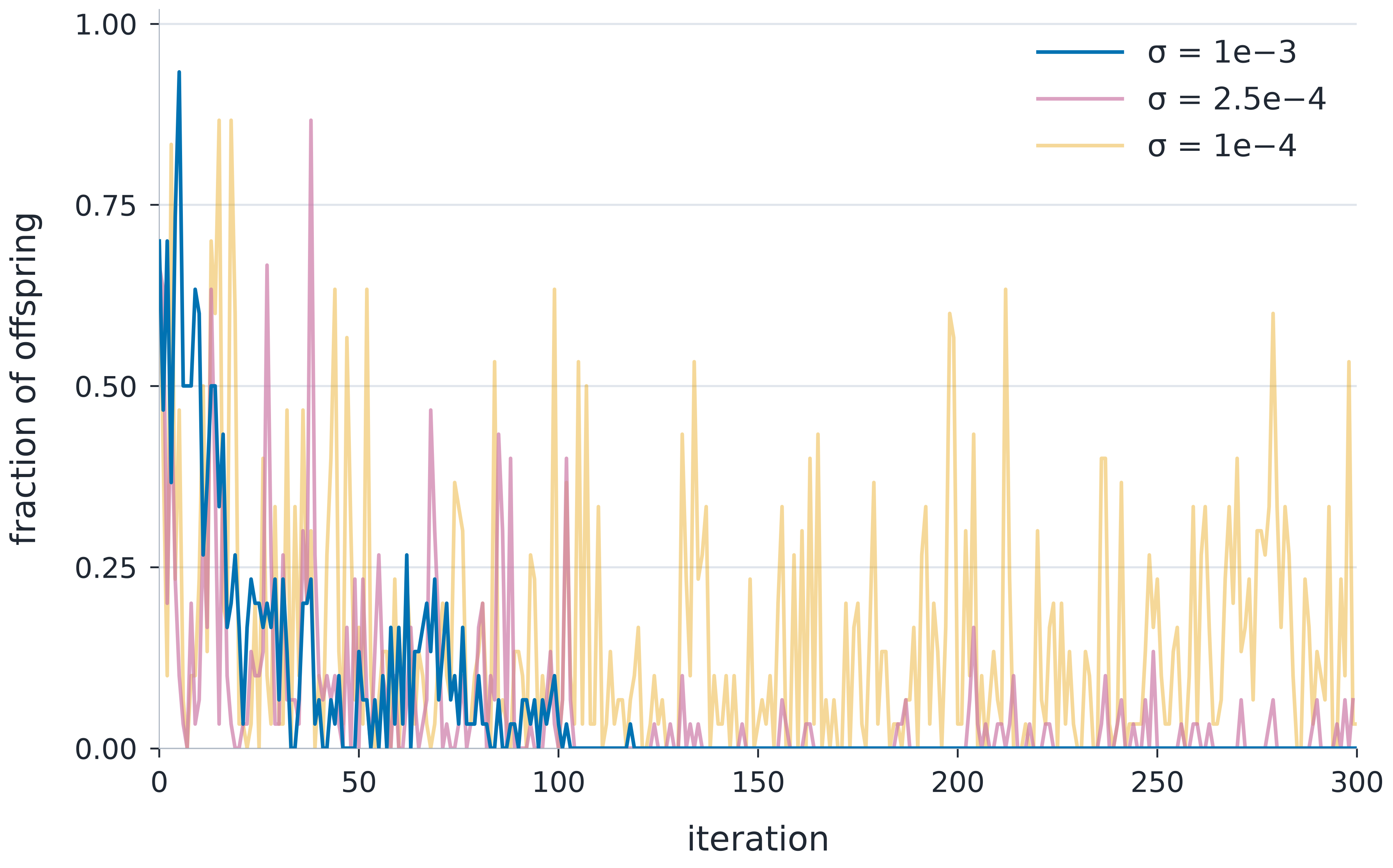}
        \caption{Fraction of  offspring that outperform parent.}
        \label{fig:small_sigma_comparison_frac}
    \end{subfigure}
    \caption{Effect of perturbation magnitude: a) A comparison in fine-tuning performance on Sudoku with perturbation magnitudes where the paradox occurs ($\sigma = 1e-3$) vs perturbation magnitudes where offspring produce viable improvement ($\sigma \in \{1e-4, 2.5e-4\}$). b) The fraction of better offspring compared to the parent under different perturbation magnitudes. Even though the smaller perturbation magnitudes produce better candidates, the larger perturbations produce better directions and thus a better overall training run.}
    \label{fig:small_sigma_comparison}
\end{figure}

If large $\sigma$ values in the range where the paradox occurs (e.g.\ $\sigma = 10^{-3}$) lead to worse performance, why not use $\sigma$ in the range of $(1.0 \times 10^{-4},2.5 \times 10^{-4})$ where improvement is possible? Indeed, in Figure~\ref{fig:magnitude_tests}, when $\sigma$ is small enough, the distribution sometimes exceeds parent performance, suggesting that opportunities for improvement exist. However, when $\sigma$ is set to these values, ES progress is significantly less, as is tested and shown in Figure~\ref{fig:small_sigma_comparison}. The improvements are lost in the noise, as discussed in detail in Section~\ref{sec:es_update}. However, even though large $\sigma$ values result in degraded performance, it turns out they can be used to identify good directions reliably, as discussed in the next section.

\section{How Are Good Directions Identified?}
\label{sec:correlation}

Section~\ref{sec:landscape} leads to the question: When all models perform worse than the current model, does their relative performance still indicate which directions are more promising? Although it is not possible to directly measure the reward gradient for model checkpoints, it is possible to evaluate the relative performance of perturbations across magnitudes where improving directions do and do not occur. If significant correlations are seen across magnitudes, regardless of performance, then relative performance is indicative of good directions regardless of all perturbation offspring performing worse than the parent. 

\begin{figure}
    \centering
    \includegraphics[width=\linewidth]{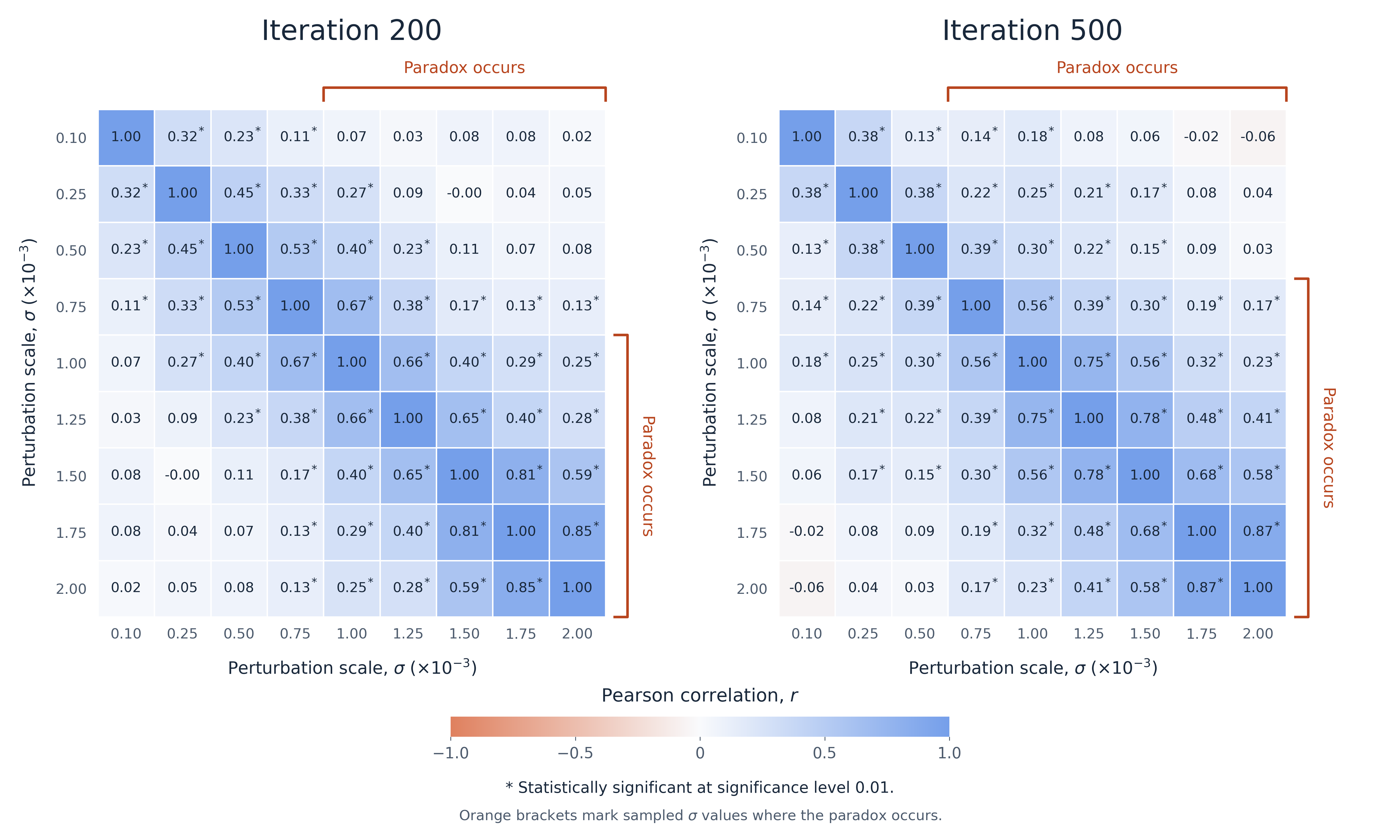}
    \caption{Correlation between rewards at various perturbation magnitudes $\sigma$ at iterations 200 and 500 iterations in the Sudoku Fine-Tuning task. The $\sigma$ range at which the paradox occurs, i.e.\ the best of the 30 offspring has a lower reward than the parent, is labeled with red arrows. The rewards at small and large $\sigma$ correlate significantly. Thus, the same good directions can be identified at various perturbation magnitudes, even at which all the perturbations are worse than the parent.}
    \label{fig:correlation_200_500}
\end{figure}

To test this hypothesis, the checkpoints from Sudoku and Countdown runs were again used. The set of 300 perturbation directions was evaluated on each task's respective 800-problem dataset at magnitudes corresponding to $\sigma = 10^{-4}$ and $\sigma$ from $2.5 \times 10^{-4}$ to $2 \times 10^{-3}$ in steps of $2.5 \times 10^{-4}$. A correlation matrix was then constructed between the relative rewards of perturbations at these magnitudes. If the perturbation rewards at large $\sigma$ contain a signal, then they should correlate with the rewards of the same perturbations at small $\sigma$. From Figure ~\ref{fig:correlation_200_500} we can see indeed, the rewards are significantly correlated between large and small perturbations, even when the rewards of large perturbations are significantly lower than those of the small perturbations. While greater magnitude differences reduce the measured correlation, the correlations remain positive in most cases.

These correlations likely emerge due to a special property of ES with isotropic Gaussian noise in high-dimensional settings. This noise occupies a thin shell in what has been referred to as a ``soap bubble'' effect \citep{huszar2017soapbubbles}. In a high-dimensional space, all perturbations are nearly identical in norm, effectively canceling out any isotropic curvature in the reward landscape, while leaving differences in relative performance due to the gradient. In other words, the magnitude of noise is identical in all directions, and thus, relative performance allows a signal to still emerge.

\begin{figure}
    \centering
    \begin{subfigure}[t]{0.48\linewidth}
        \centering
        \includegraphics[width=\linewidth]{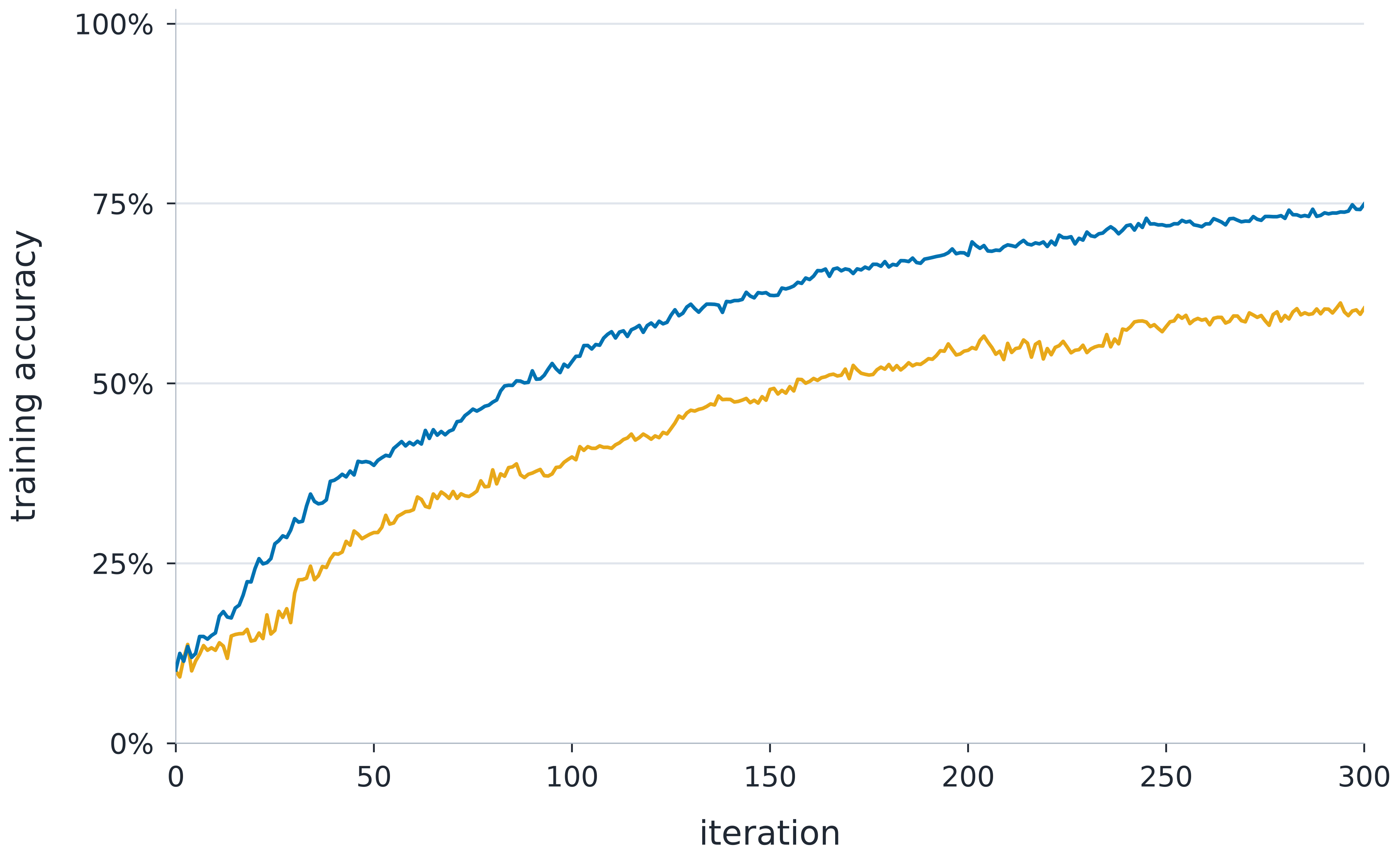}
        \caption{Training performance of scaled best-of-30 perturbation selection compared with the actual aggregated ES update.}
        \label{fig:bestof_30_train}
    \end{subfigure}\hfill
    \begin{subfigure}[t]{0.48\linewidth}
        \centering
        \includegraphics[width=\linewidth]{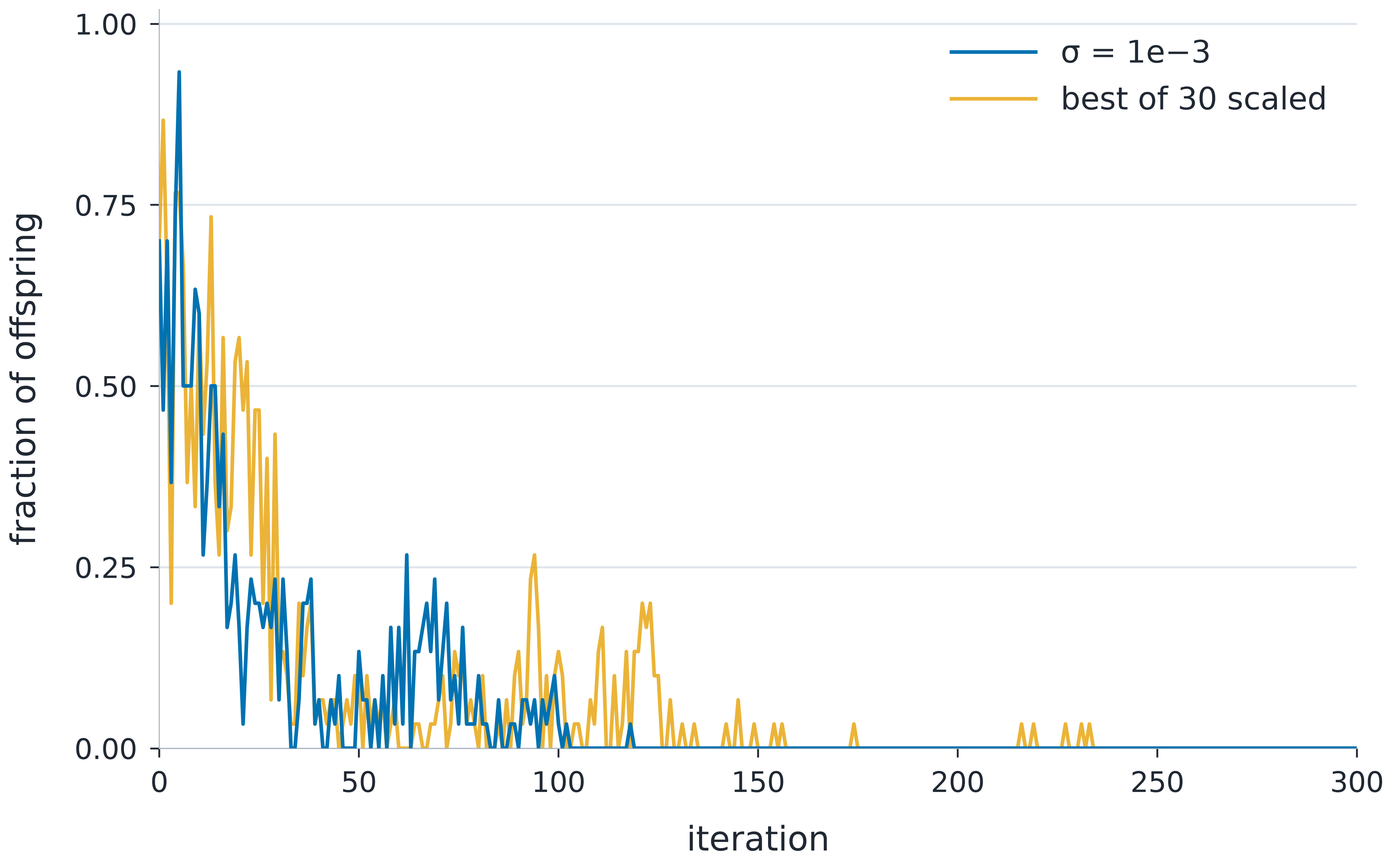}
        \caption{Fraction of the perturbations with the best-of-30 method that outperform the parent.}
        \label{fig:bestof_30_frac}
    \end{subfigure}
    \caption{Best-of-30 selection ablation: (a) Training performance of scaled best-of-30 perturbation selection compared with the actual aggregated ES update. (b) Fraction of the perturbations with the best-of-30 method that are better than the parent. When scaled from a size where the best perturbation performs worse to the size of a typical ES update, the model is able to improve. However, the aggregated ES update direction still significantly outperforms this best-of-30 direction, suggesting that aggregation contributes further to performance.}
    \label{fig:bestof_30}
\end{figure}

One way to take advantage of the good directions at large $\sigma$ would be to simply follow the best perturbation, instead of performing the aggregation through a weighted sum. To evaluate this ablation, a second test was implemented. The perturbation rewards were sampled on the standard (large) $\sigma$ (1e-3), the best of them identified, and the perturbation was scaled down to the size of an average ES update (i.e.\ a small $\sigma$). A 500-iteration training run was then run with this process. The training curves along with the fraction of perturbations better than the parent are displayed in Figure~\ref{fig:bestof_30}. While this best-of-N selection allows making progress, it is still outperformed by the actual ES update. This result suggests that an additional mechanism of aggregation contributes further to the ES performance, as will be discussed next.

%\section{An Analytical Model of Evolution Strategies}
\section{How Do the Updates Improve Performance?}
\label{sec:es_update}

Sections~\ref{sec:landscape} and~\ref{sec:correlation} demonstrated that model performance decreases with increasing perturbation; however, even when all offspring are worse than the parent, good search directions can be identified among them.  This section demonstrates how the ES aggregation mechanism adds value. The insights are drawn from an analytical model of the ES update in a reward landscape with observation noise and quadratic loss. In this model, the ES update can be decomposed into a \emph{signal} component along the gradient and a \emph{noise} component orthogonal to it. The model shows that, in expectation, the ES update moves a fixed distance in the direction of the signal, while reducing the magnitude of the noise component via aggregation of offspring. The mean and variance of the ES update in this model were derived independently by \citep{hoy2026matchingaccuracydifferentgeometry}, along with a similar decomposition of signal and noise of the update, which they use to estimate parameter drift over training runs. The analysis here differs by focusing on the per-step reward improvement and its relationship to observation noise and curvature.  

In the subsections below, the update decomposition is first defined in a simplified linear reward landscape. A realistic observation noise term is then introduced, and the model is extended to include curvature. In the end, the optimal perturbation magnitude is determined in the model, showing that it results in likely improvements. All along, these results are validated empirically with existing training runs.

\subsection{Defining the Signal and the Noise Components of the ES Update}

Recall the definition of ES update in \Eqref{eq:es-update}.
Let $g = \nabla R(\theta_t)$ be the gradient in the reward landscape and $\hat g = g/\|g\|$ its normalized direction. Each perturbation $\epsilon_i$ can be divided into a signal component along the gradient, $u_i\,\hat g$, and a perpendicular noise component, $\epsilon_i^\perp$, as
\begin{equation}
    \epsilon_i \;=\; u_i\,\hat g \;+\; \epsilon_{\perp i},
    \qquad
    u_i = \hat g^{\!\top}\epsilon_i,
    \qquad
    \label{eq:eps-split-paper}
\end{equation}

where $\epsilon_i^\perp \sim \mathcal{N}\!\big(0,\, I_d - \hat g\hat g^{\!\top}\big)$ and $u_i \sim \mathcal{N}(0,1)$ (see Appendix~\ref{sec:es-signal-noise} for details). This decomposition can then be applied to the ES update equation \Eqref{eq:es-update}, dividing it into an update along the gradient and an update perpendicular to the gradient:

\begin{equation}
    \Delta\theta_{\parallel} \;=\; \Big(\frac{\alpha}{N}\sum_{i=1}^N Z_iu_i\hat g\Big),
    \qquad
    \Delta\theta_\perp \;=\; \frac{\alpha}{N}\sum_{i=1}^N Z_i\,\epsilon_{\perp i} .
    \label{eq:par-perp-paper}
\end{equation}
 
Under linear reward, $R(\theta_i) = R(\theta_t) + g^{\!\top}(\theta_i-\theta_t)$. Then $R_i = R(\theta_t) + \sigma\|g\|\,u_i$. Using an assumption that the sampled reward mean and standard deviation match their expected values, $\mu_R = R(\theta_t)$ and $\sigma_R = \sigma\|g\|$, and an individual perturbation's z-score is
\begin{equation}
    Z_i \;=\; \frac{\sigma\|g\|\,u_i}{\sigma\|g\|} \;=\; u_i .
    \label{eq:Z-noiseless-paper}
\end{equation}

Each individual perturbation, as weighted in the ES update, can be written as its signal and noise components. The weighted signal component of perturbation $i$ becomes $\alpha u_i^2 \hat g$, with $u_i^2  \sim \chi^2_1$, while the weighted noise component becomes $u_i\epsilon_{\perp i}$, allowing for their expected norm and squared norm of scaled perturbation signal and noise components to be calculated as

\begin{align}
    \text{perturbation signal:}\qquad \mathbb{E}[\|\alpha u_i^2 \hat g\|] &\;=\; \alpha,
    \label{eq:perturbation-signal-noiseless}\\
    \text{perturbation noise:}\qquad \sqrt{\mathbb{E}\left[\|u_i\epsilon_{\perp i}\|^2\right]} &\;=\; \alpha\sqrt{d-1} .
    \label{eq:perturbation-noise-noiseless}
\end{align}

When switching to the full ES update aggregation, $\sum_iu_i^2 \sim \chi^2_N$, and the update magnitudes are

\begin{align}
    \text{update signal:}\qquad \mathbb{E}[\hat g^{\!\top}\Delta\theta] &\;=\; \alpha,
    \label{eq:signal-noiseles-paper}\\
    \text{update noise:}\qquad \sqrt{\mathbb{E}\left[\|\Delta\theta_\perp\|^2\right]} &\;=\; \alpha\sqrt{\frac{d-1}{N}} .
    \label{eq:noise-noiseless-paper}
\end{align}

Notice that the expected magnitude of the update component along the gradient is the same as the magnitude of the perturbation component on the gradient, while the update noise component shrinks with $\sqrt{N}$. This result shows that the aggregation of the ES update maintains the expected signal constant while reducing noise. For full derivations, see Appendix~\ref{sec:es-signal-noise}.

\subsection{Including Observation Noise}

\begin{figure}
    \centering
    \includegraphics[width=1\linewidth]{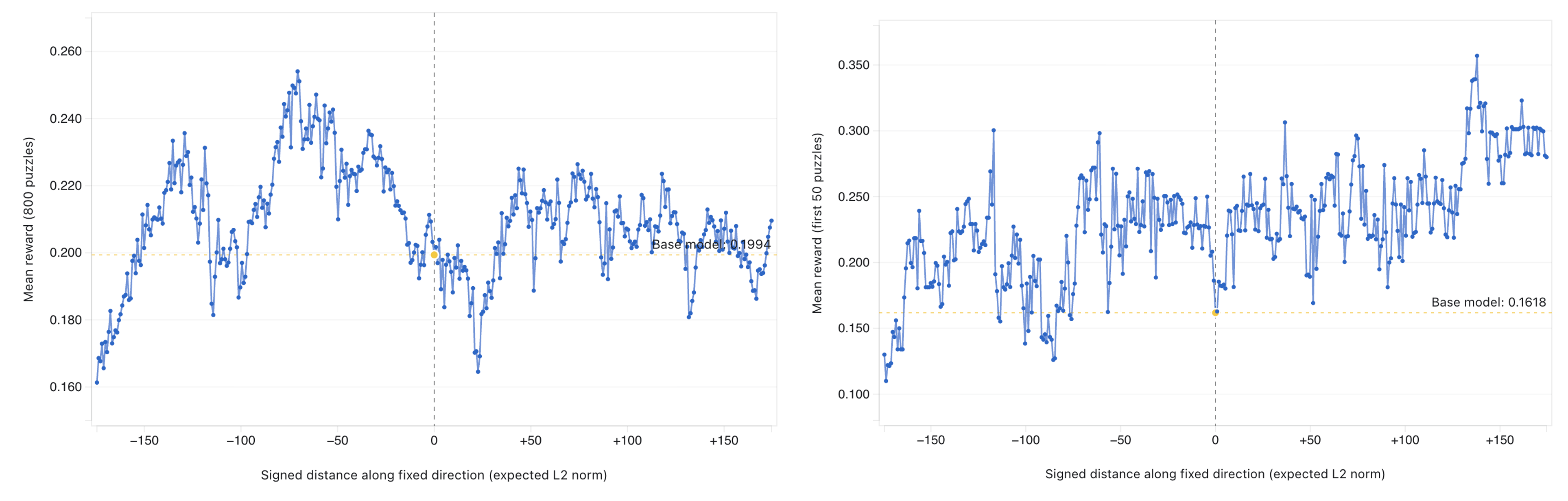}
    \caption{Average accuracy of Qwen 2.5 7B on Sudoku task along a random direction from $-2\sigma$ to $+2\sigma$. Left: Landscape on a set of 800 problems. Right: Landscape on a set of 50 problems. A significant portion of the variance occurs at frequencies much higher than the perturbation size used in ES, and can thus be treated as observation noise.}
    \label{fig:random_direction}
\end{figure}

In the ES fine-tuning tasks, the models are evaluated at temperature zero, and the model outputs are therefore deterministic. However, when exploring along specified directions, noise occurs in the reward landscape at much higher frequencies than the magnitudes of the perturbations used (as shown in Figure~\ref{fig:random_direction}).Similar to the nugget term in Gaussian Process models \citep{pepelyshev2010rolenuggettermgaussian}, this noise can be incorporated in the analytical model as observation noise at scale $\sigma_\xi$:

\begin{equation}
    R_i \;=\; R(\theta_t + \sigma\epsilon_i) \;+\; \sigma_\xi\, \xi_i,
    \qquad \xi_i \sim \mathcal{N}(0,1).
    \label{eq:obs-model-paper}
\end{equation}

Plugging this reward function into the ES update (\Eqref{eq:es-update}) produces the expected signal, noise, and reward improvement as

\begin{equation}
    \mathbb{E}[\hat g^{\!\top}\Delta\theta]
    = \frac{\alpha\sigma\|g\|}{\sqrt{\sigma^2\|g\|^2 + \sigma_\xi^2}},
    \quad
    \sqrt{\mathbb{E}\left[\|\Delta\theta_\perp\|^2\right]}
    = \alpha\sqrt{\frac{d-1}{N}},
    \quad
    \mathbb{E}[\Delta R]
    = \frac{\alpha\sigma\|g\|^2}{\sqrt{\sigma^2\|g\|^2+\sigma_\xi^2}}.
    \label{eq:noisy}
\end{equation}

Notice that with observation noise, both the expected signal and reward increase with increasing $\sigma$. Thus, whereas the variance in the reward for small $\sigma$ is mostly produced by noise, larger $\sigma$ increases reward variance, making the ES estimate more accurate. For full derivations, see Appendix~\ref{sec:es-signal-noise}.

\subsection{Including Landscape Curvature}

In the third step, the analytical model is extended to include general downward curvature $Q$. The reward can then be written as
\begin{equation}
R(\theta) = R(\theta_t) + g^{\!\top}(\theta-\theta_t) - \tfrac12(\theta-\theta_t)^{\!\top}Q\,(\theta-\theta_t).
\end{equation}
The expected signal, noise, and reward are
\begin{equation}
    \text{signal:}\qquad
    \mathbb{E}[\hat g^{\!\top}\Delta\theta]
    \;=\;
    \frac{\alpha\sigma\|g\|}{\sigma_R}
    \;=\;
    \frac{\alpha\|g\|}{\sqrt{\|g\|^2 \;+\; \dfrac{\sigma^2}{2}\operatorname{tr}[Q^2] \;+\; \dfrac{\sigma_\xi^2}{\sigma^2}}} ,
    \label{eq:signal-quad-paper}
\end{equation}
\begin{equation}
    \text{noise:}\qquad
    \sqrt{\mathbb{E}\left[\|\Delta\theta_\perp\|^2\right]}
    \;\approx\;
    \alpha\sqrt{\frac{d-1}{N}},
    \label{eq:noise-quad-paper}
\end{equation}
\begin{equation}
    \text{reward:}\qquad
    \mathbb{E}[\Delta R]
    \;\approx\;
    \underbrace{\frac{\alpha\|g\|^2}{\sqrt{\|g\|^2 \;+\; \dfrac{\sigma^2}{2}\operatorname{tr}[Q^2] \;+\; \dfrac{\sigma_\xi^2}{\sigma^2}}}}_{\text{signal gain}}
    \;-\;
    \underbrace{\frac{\alpha^2}{2N}\operatorname{tr}[Q]}_{\text{noise penalty}},
    \label{eq:dR-quad-main}
\end{equation}
where the terms that are negligible for noise and reward have been dropped. 
The crucial property identified in the linear case also holds with observation noise and curvature: the signal component of the update is the same for a perturbation as it is for the aggregate update, but the noise component of the aggregate is smaller. 

With curvature, a tradeoff occurs between increasing perturbation size to avoid observation noise $\sigma_\xi^2$ and decreasing the perturbation size to reduce the effect of curvature via $\operatorname{tr}[Q^2]$. This tradeoff has been documented with perturbation-based methods before \citep{multivariate_stochastic_approx_perturbation,duchi2015optimal,nesterov2017random}. However, in high dimensions, the ES estimate has two unique properties. First, population size only affects the magnitude of the noise component. Thus, the expected gain in the reward plateaus once the population size reduces noise enough so that the ES update is no longer affected by curvature. Second, while the gain in reward plateaus with additional population members, the magnitude of update itself falls proportional to $\frac{1}{\sqrt{N}}$.

\begin{figure}[t]
    \centering
    \begin{subfigure}[t]{0.48\linewidth}
        \centering
        \includegraphics[width=\linewidth]{\detokenize{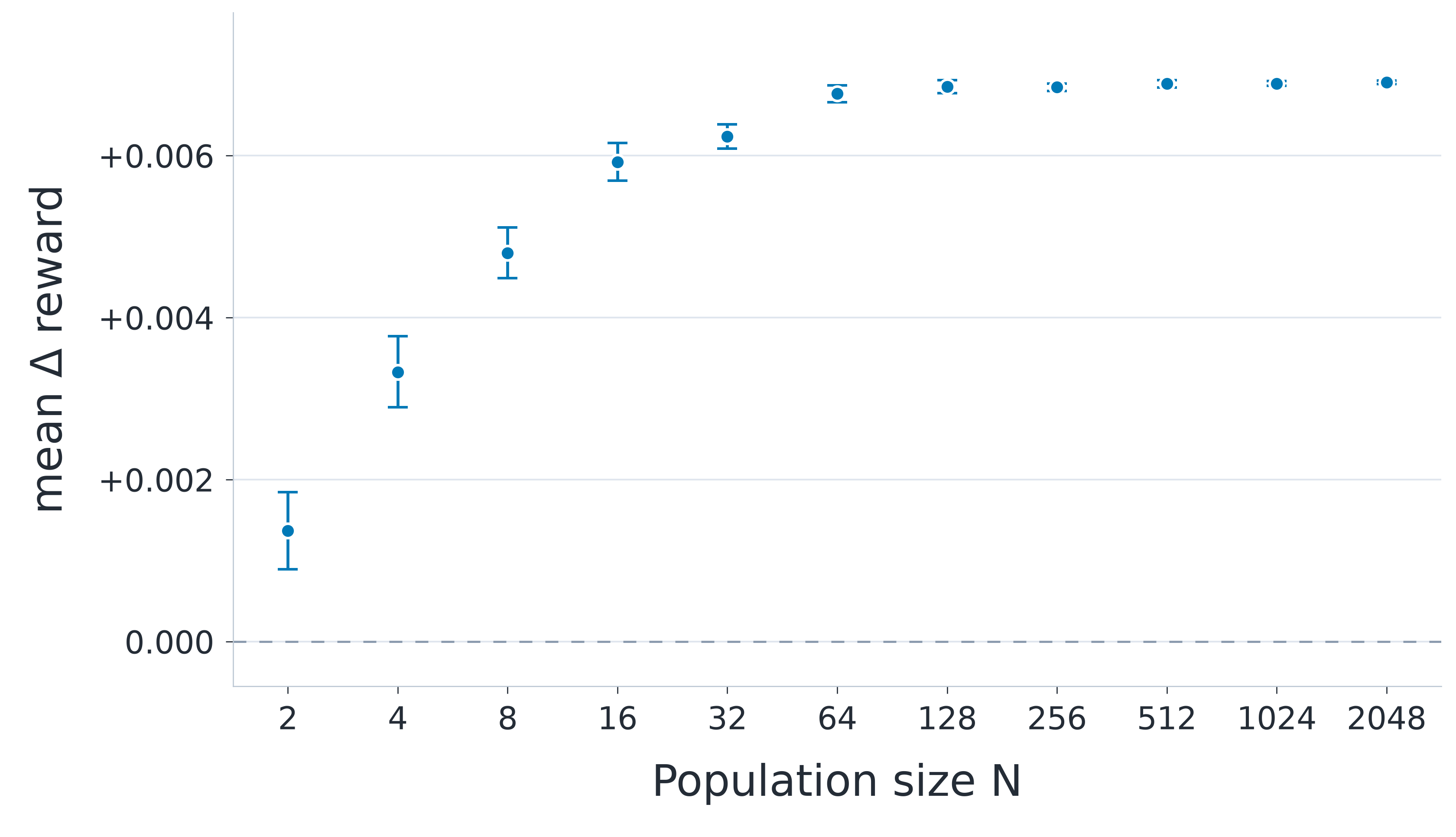}}
        \caption{MNIST: Reward improvement vs. population size}
        \label{fig:mnist_population_tests_a}
    \end{subfigure}\hfill
    \begin{subfigure}[t]{0.48\linewidth}
        \centering
        \includegraphics[width=\linewidth]{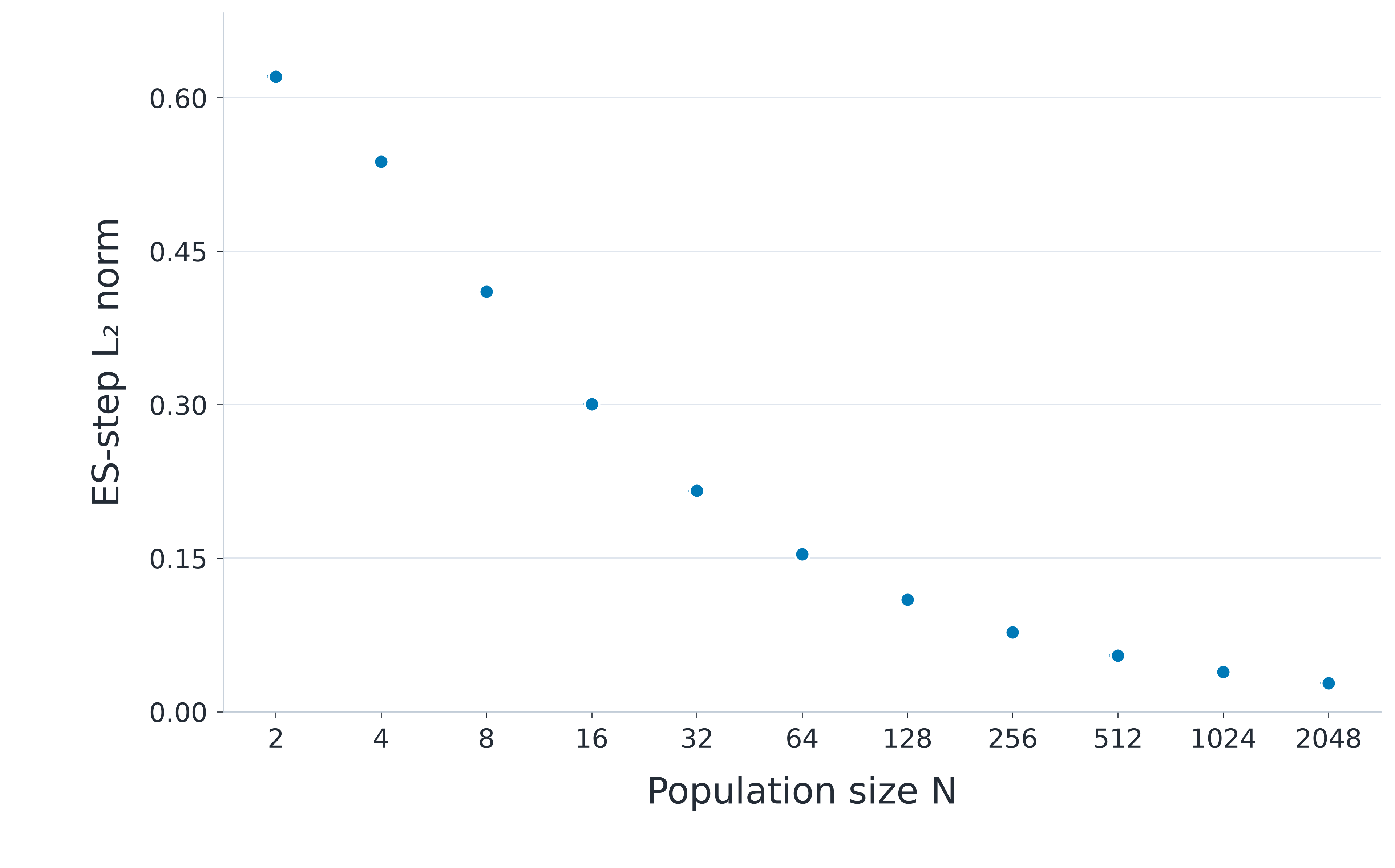}
        \caption{MNIST: Update size vs. population size}
        \label{fig:mnist_population_tests_b}
    \end{subfigure}

    \vspace{0.5em}

    \begin{subfigure}[t]{0.48\linewidth}
        \centering
        \includegraphics[width=\linewidth]{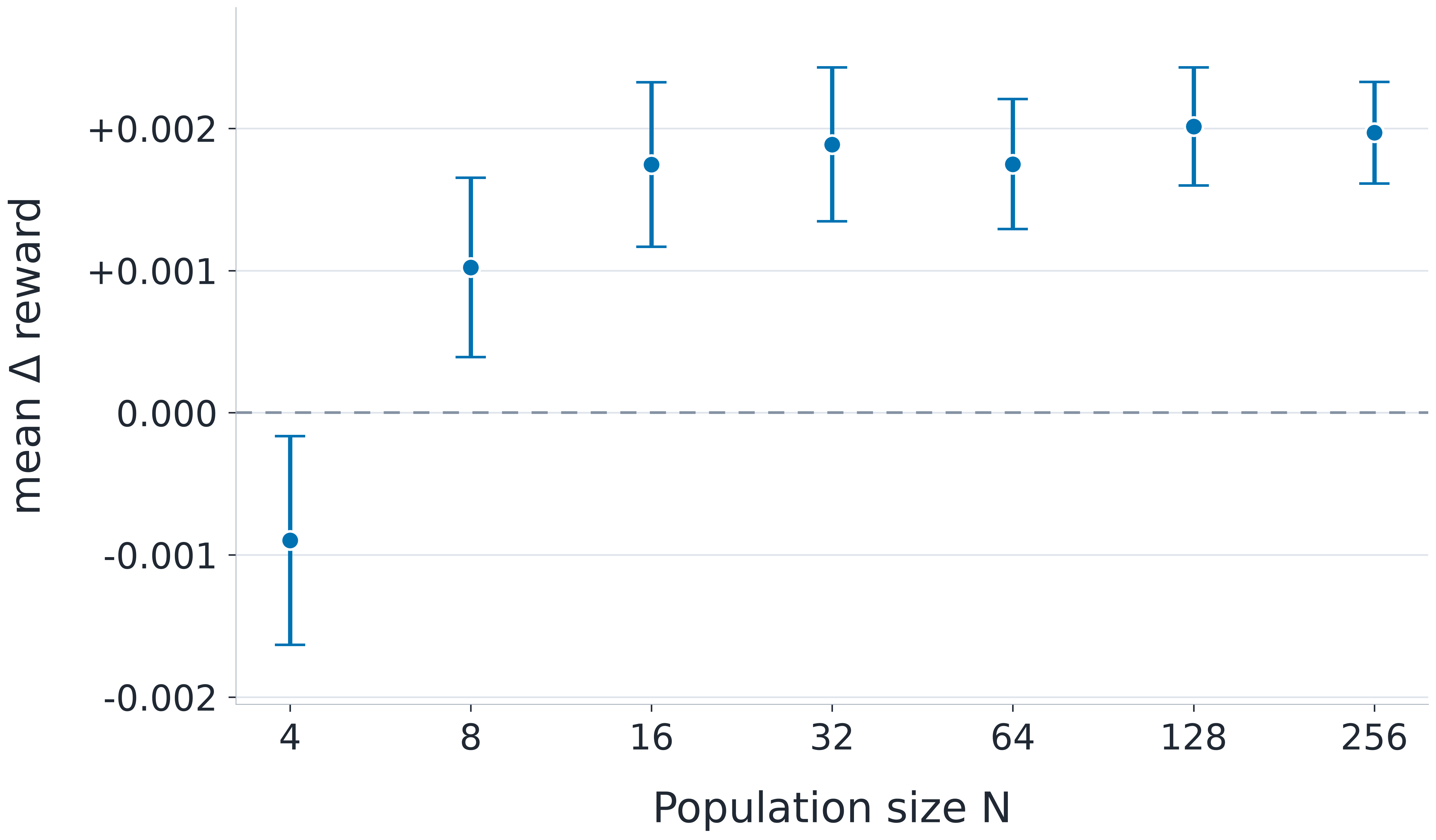}
        \caption{Fine-tuning: Reward improvement vs. population size}
        \label{fig:fine_tuning_update_size_magnitudes_a}
    \end{subfigure}\hfill
    \begin{subfigure}[t]{0.48\linewidth}
        \centering
        \includegraphics[width=\linewidth]{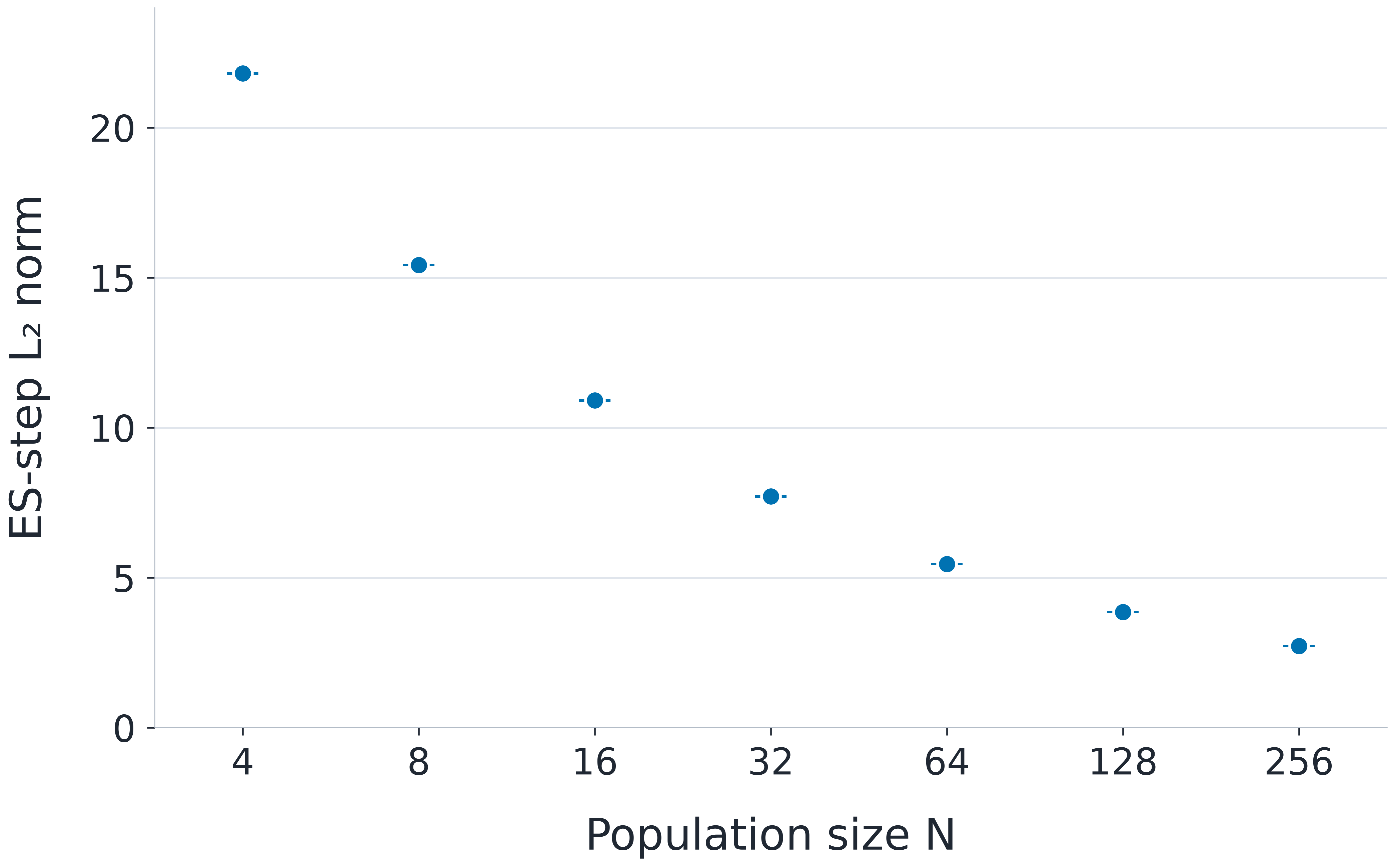}
        \caption{Fine-tuning: Update size vs. population size}
        \label{fig:fine_tuning_update_size_magnitudes_b}
    \end{subfigure}

    \caption{Update reward improvement and update magnitude vs. population size. (a) Distributions of reward improvements from ES updates for different population sizes for CNN on MNIST. (b) Magnitude of ES updates in MNIST. (c) Mean and confidence intervals of improvement from ES updates at different population sizes for the Sudoku LLM fine-tuning task. (d) Magnitude of ES update for various population sizes. Distributions were wide, so for clarity, only means and CIs are shown. The model was Qwen 2.5 7B. In both settings, update reward improvement increases with population size and then plateaus, while the norm of the ES update decreases with population size.}
    \label{fig:update_size_magnitudes}
\end{figure}

To test whether these properties show up empirically with ES optimization, checkpoints were taken every 100 iterations from the Sudoku 500 iteration fine-tuning run, as well as from the MNIST training run. With each checkpoint, 30 ES updates were drawn from independent sets of $N$ perturbations, where $N$ is the designated population size. The reward of these updates was measured, and confidence intervals drawn for the mean reward update (Figure~\ref{fig:update_size_magnitudes}). Notice that, as predicted by the noise-curvature model, the expected reward update increases and then plateaus with population size. Further, the update sizes are highly consistent, and even as their magnitude decreases significantly with respect to population size, the reward improvement remains the same. This result demonstrates that the ES update is holding the signal constant while reducing noise.

These plots also illuminate why ES is able to work on billion-parameter models with population sizes of only 30: once the noise is reduced enough to avoid the curvature penalty, ES achieves its maximum possible improvement on each step.

\begin{figure}
    \centering
    \begin{subfigure}[t]{0.48\linewidth}
        \centering
        \includegraphics[width=\linewidth]{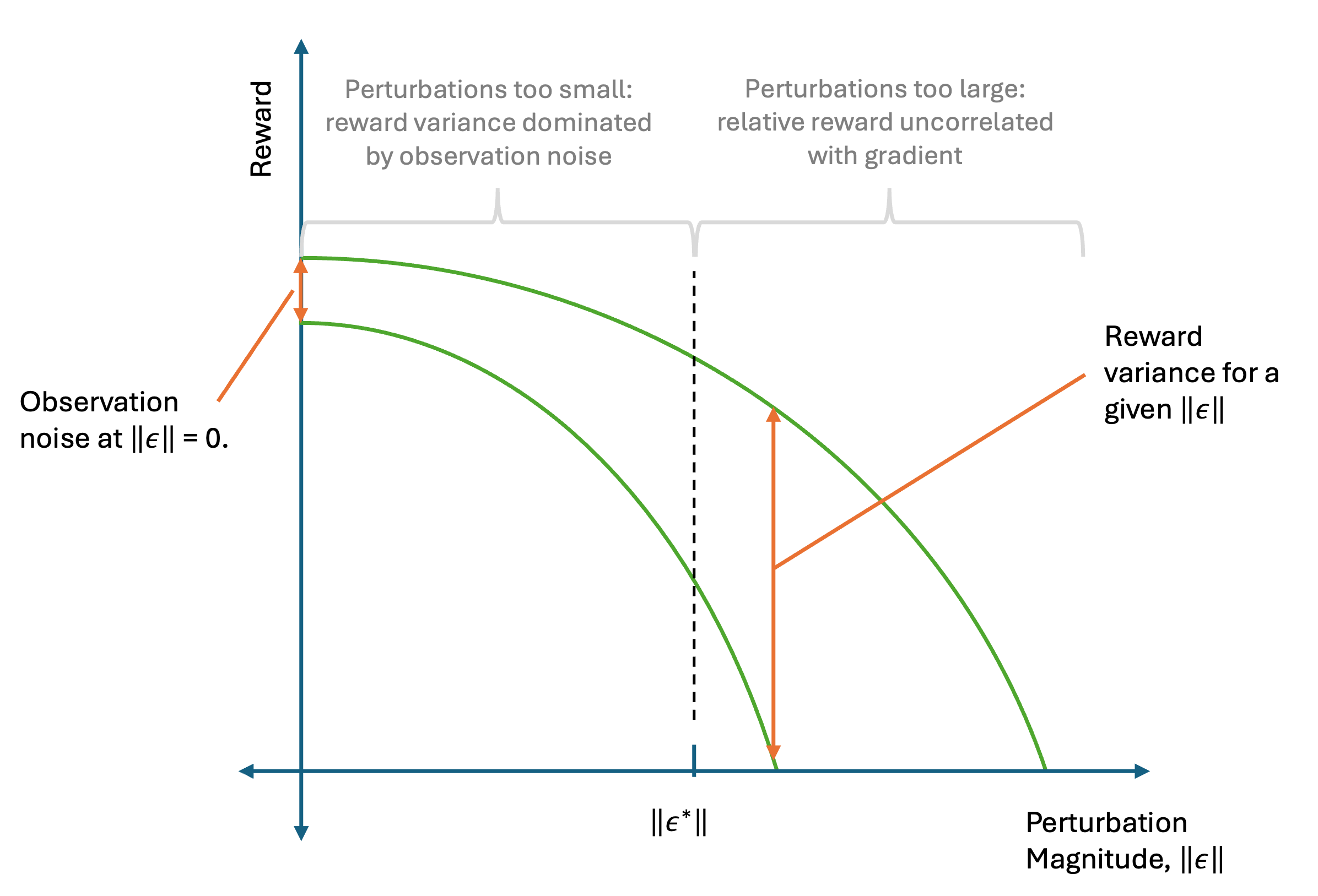}
        \caption{Single Step Reward vs.\ perturbation}
        \label{fig:single_step_reward_vs_perturbation}
    \end{subfigure}\hfill
    \begin{subfigure}[t]{0.48\linewidth}
        \centering
        \includegraphics[width=\linewidth]{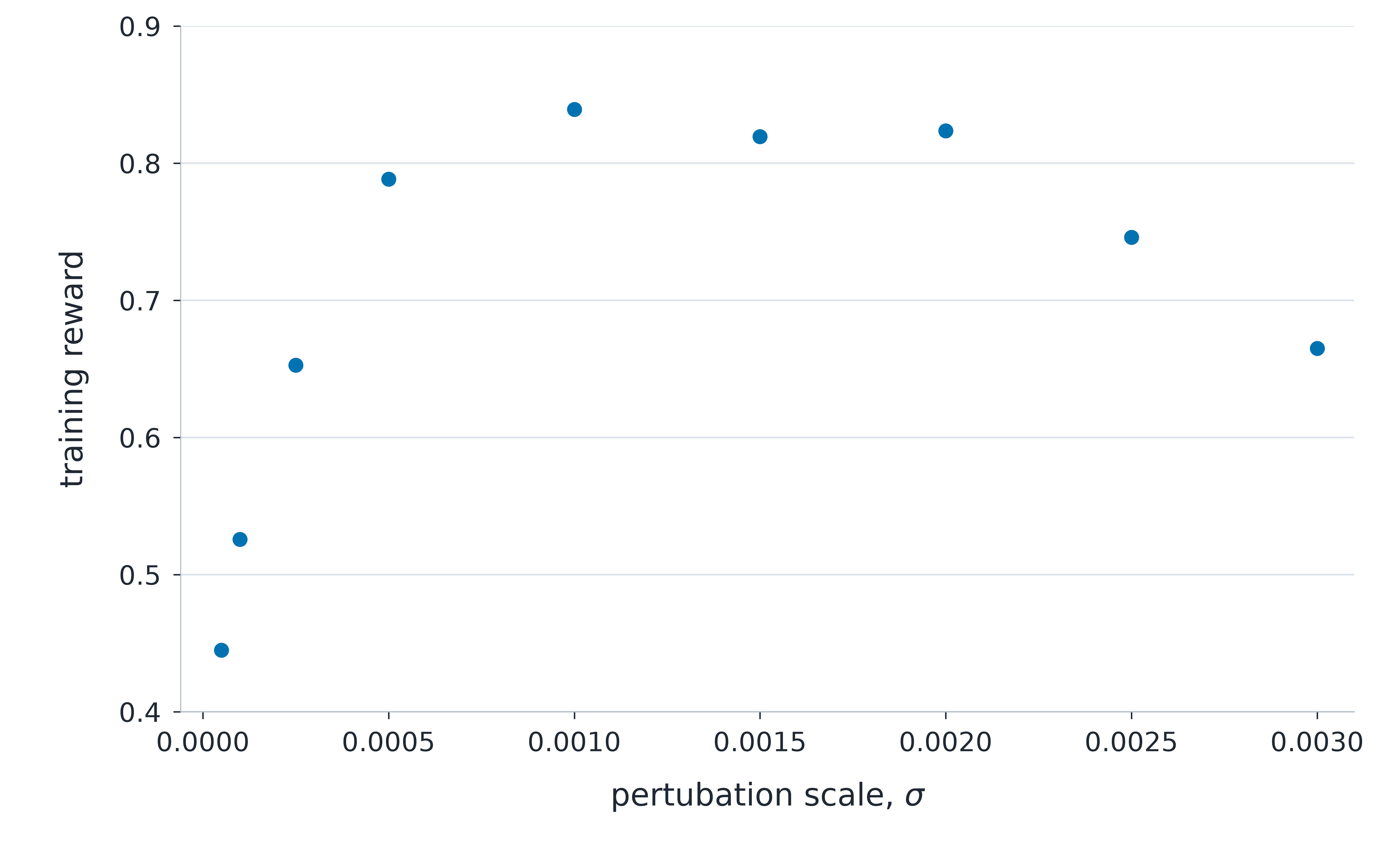}
        \caption{Training Performance vs. $\sigma$}
        \label{fig:training_performance_vs_sigma}
    \end{subfigure}
    \caption{(a): A visual representation of reward vs.\ perturbation magnitude, demonstrating why the offspring are worse than the parent. Perturbations happen at a fixed size; if they are too small, observation noise drowns out differences between good and bad directions, and if they are too big, the correlation between perturbation reward and direction quality breaks down. The paradox occurs when the magnitude is at a distance where all perturbations have a reward worse than the current model, but they still differentiate the good directions from the bad. (b) A demonstration of the tradeoff in perturbation size empirically, with performance of Qwen 2.5 7B on the Sudoku task after 300 ES iterations varying $\sigma$, holding all else constant. Results illustrate observation noise deteriorating performance at small values of $\sigma$, and correlations breaking down at large values, demonstrating the tradeoff visualized by (a).}
    \label{fig:sigma_graphic}
\end{figure}

\subsection{Effective Perturbation Scale Triggers The Paradox}
\label{sec:optimal-sigma}

As shown in Section~\ref{sec:landscape}, testing an alternative $\sigma$ value that avoids the paradox revealed that perturbation magnitudes that invoked the paradox outperformed smaller perturbations that did not. The noise-curvature model offers a tradeoff between observation noise and curvature. If $\alpha$, $g$, $Q$ and $\sigma_\xi$ are fixed, the expected improvement in \Eqref{eq:dR-quad-main} is maximized by minimizing $f(\sigma^2) = \sigma_\xi^2/\sigma^2 + \tfrac{\sigma^2}{2}\operatorname{tr}[Q^2]$ over $\sigma^2$. This setting gives
\begin{equation}
    \sigma_*^4 \;=\; \frac{2\,\sigma_\xi^2}{\operatorname{tr}[Q^2]},
    \qquad
    \sigma_* \;=\; \left(\frac{2\,\sigma_\xi^2}{\operatorname{tr}[Q^2]}\right)^{1/4}.
    \label{eq:sigma-opt-paper}
\end{equation}
 
To characterize the population at this optimal sigma, using the fact that $\mathbb{E}[\epsilon_i^{\!\top}Q\epsilon_i] = \operatorname{tr}[Q]$, the expected difference in the reward from the parent is
\begin{equation}
    \mu_R - R(\theta_t) \;=\; -\frac{\sigma^2}{2}\operatorname{tr}[Q] ,
    \qquad
    \mu_R - R(\theta_t)\Big|_{\sigma_*} \;=\; -\,\sigma_\xi\sqrt{\frac{r_{\mathrm{eff}}}{2}},
    \qquad
    r_{\mathrm{eff}} := \frac{(\operatorname{tr}[Q])^2}{\operatorname{tr}[Q^2]} ,
    \label{eq:mean-gap-paper}
\end{equation}
where $r_{\mathrm{eff}} \in [1, d]$ is the participation ratio (effective rank) of the Hessian, equal to $d$ for isotropic curvature and to $r$ for a rank-$r$ Hessian with equal eigenvalues. In units of the population spread,

\begin{equation}
    \frac{R(\theta_t) - \mu_R}{\sigma_R}%\Big|_{\sigma_*}
    \;=\;
    \frac{\frac{\sigma^2}{2}\operatorname{tr}[Q]}{\sqrt{\sigma^2\|g\|^2 \;+\; \frac{\sigma^4}{2}\operatorname{tr}[Q^2] \;+\; \sigma_\xi^2}}
    \;=\;
    \frac{\operatorname{tr}[Q]}{2\sqrt{\frac{1}{\sigma^2}\|g\|^2 \;+\; \frac{1}{2}\operatorname{tr}[Q^2] \;+\; \frac{\sigma_\xi^2}{\sigma^4}}},
    \label{eq:gap-sigmas-paper}
\end{equation}

where $\sigma_R$ is the standard deviation of the population rewards. \Eqref{eq:gap-sigmas-paper} quantifies how the rewards of perturbations distribute compared to the parent reward $R(\theta_t)$. A larger value of \Eqref{eq:gap-sigmas-paper} corresponds to a higher probability for the perturbation to be worse than the parent. 

Since the value of \Eqref{eq:gap-sigmas-paper} is increasing with respect to $\sigma$, a smaller $\sigma$ decreases the probability of generating worse populations. In the meantime, a smaller $\sigma$ ($\ll\sigma_*$) also reduces the expected improvement of ES updates, according to \Eqref{eq:dR-quad-main}. As a result, a smaller $\sigma$ decreases the chance for the paradox to happen. For larger $\sigma$, although the chance for producing worse perturbations is increased, the expected improvement is also reduced (if $\gg\sigma_*$), resulting in fewer occurrences of the paradox. This is demonstrated visually in Figure~\ref{fig:single_step_reward_vs_perturbation} and empirically in Figure~\ref{fig:training_performance_vs_sigma}, which shows that if a $\sigma$ value provides a larger improvement for ES updates, its chance to trigger the paradox also increases. Therefore, the paradox is a byproduct of a highly effective ES run.

\section{Discussion and Future Work}

The ES paradox emerged as an unexpected anomaly in population-based search in large spaces. Such methods are presumed to work by creating offspring that are better than parents, yet in many cases ES is able to consistently improve even when this is not true. Understanding the mechanisms behind this paradox led to a new way of thinking about search in high-dimensional spaces: instead of an anomaly, the paradox is a side effect of a principled interaction of noise and curvature in such spaces. This mechanism is different from gradient descent, simulated annealing, and other evolutionary computation, so much so that it can be seen as a paradigm shift in the sense of \citet{kuhn:revolutions}: an anomaly becomes a fundamental principle in the new way of seeing things. As a result, old assumptions need to be re-evaluated, and new opportunities arise for progress. For instance:

\paragraph{Small Population Performance of ES:}
One of the most dramatic features of the new line of ES work in LLM fine-tuning has been the significantly smaller population size. Previously, it had been assumed that ES had fundamental scaling limitations with respect to high-dimensional spaces \citep{jamieson2012query}, yet within fine-tuning, it has been competitive with, and often even outperforming, gradient-based reinforcement learning methods. This work now sheds some light on this phenomenon, demonstrating that once the noise component of the ES update is small enough to avoid incurring a significant cost from curvature, the update is able to achieve the maximum improvement in reward - even if the noise component is orders of magnitude larger than the signal component.

\paragraph{The surprising Robustness of LLMs to Noise in Parameter Space:}
From the update reward improvement estimate results in Figure~\ref{fig:update_size_magnitudes}, per step update is seen to plateau at approximately a population size of 30. Given the analysis in Section~\ref{sec:es_update}, this would imply each ES update contains a signal component of maximum magnitude $\alpha$, while the noise component contains a magnitude of $\alpha\sqrt{\frac{d-1}{N}} = \alpha\sqrt{\frac{7\times10^{9}}{30}} \approx \alpha \times 15,000$. While fine-tuning, for every unit distance ES moves the network in the direction of model improvement, it moves 15,000 unit distances in an irrelevant direction. This is consistent with  \citep{hoy2026matchingaccuracydifferentgeometry} which finds that the total parameter drift of ES-fine-tuned LLMs follows the random-walk prediction $\alpha^2 T d/N$ to within a few percent. This demonstrates a surprising fact that LLMs can have their parameters dramatically modified in random directions without affecting performance.

\paragraph{Future Work:} Results and the analytical model suggest that there may be ways to further optimize the hyperparameters of ES. Section \ref{sec:optimal-sigma} presented an optimal perturbation size conditional on observation noise and curvature. Future work may look at how to identify these properties during a run, or whether estimation of said optimums is too noisy to do online, but may be aggregated across fine-tuning runs. Additional work could also modify the ES algorithm itself to take advantage of this new paradigm, where improving offspring is no longer necessary for population-based optimization. 

\section{Conclusion}

The ES paradox presents a counterintuitive conclusion: that ES can learn while consistently generating worse offspring than its parent. This paper demonstrates why this is possible: even when all offspring perform worse than their parent, their relative performance contains valuable information for improving directions. The ES update then aggregates this signal while reducing the noise to a point where improvement can be made.
Together, these results suggest that ES in large-dimensional spaces can be seen as a new optimization paradigm: it takes advantage of the interactions of noise and curvature in such spaces the way other standard methods do not. The results provide insight into how ES is able to function effectively on billion-parameter LLMs, and suggest opportunities for improving the method further.

\newpage
\subsection*{AI use statement}

In this work, we used generative AI tools to help implement methods and provide critical ingredients for proving mathematical claims. We have reviewed all AI-assisted work. We take responsibility for the final content of this work,
including text, claims, or artifacts produced with the aid of generative AI.

\bibliography{iclr2026_conference}
\bibliographystyle{iclr2027_conference}

\appendix

\section{Signal, Noise, and the Perturbation Scale in z-Scored Evolution Strategies}
\label{sec:es-signal-noise}
 
This section starts with the basic properties of the ES estimator with z-score normalized rewards \citep{qiu2026evolutionstrategiesscalellm}, under linear and quadratic reward landscapes. The mean and covariance of the update in Sections~\ref{sec:linear-noiseless}--\ref{sec:quadratic} agree with those derived independently by \citep{hoy2026matchingaccuracydifferentgeometry}. They are restated here for completeness, in the signal/noise form used in the main text. The new material is the expected reward change in \eqref{eq:dR-quad}, the optimal perturbation scale and the population statistics at that scale (Section~\ref{sec:optimal-sigma-appendix}), and the concentration of the update norm (Appendix~\ref{app:update-norm}).
 
\subsection{The z-scored ES update}
\label{sec:es-update}
 
At iteration $t$, ES draws $N$ perturbation directions $\epsilon_i \sim \mathcal{N}(0, I_d)$, evaluates the perturbed models $\theta_t + \sigma\epsilon_i$ to obtain rewards $R_i$, and moves the parent by the reward-weighted average of the perturbations,
\begin{equation}
    \Delta\theta \;=\; \theta_{t+1} - \theta_t \;=\; \frac{\alpha}{N}\sum_{i=1}^N Z_i\, \epsilon_i ,
    \label{eq:es-update_appendix}
\end{equation}
where $\alpha$ is the step size and the weights $Z_i$ are the population-standardized rewards,
\begin{equation}
    Z_i \;=\; \frac{R_i - \mu_R}{\sigma_R},
    \qquad
    \mu_R = \frac{1}{N}\sum_{j=1}^N R_j,
    \qquad
    \sigma_R^2 = \frac{1}{N}\sum_{j=1}^N (R_j - \mu_R)^2 .
    \label{eq:zscore}
\end{equation}
The sample statistics $(\mu_R,\sigma_R)$ in \eqref{eq:zscore} are replaced by their population counterparts, $\mu_R = \mathbb{E}[R_i]$ and $\sigma_R^2 = \operatorname{Var}[R_i]$, in the derivations below.
 
\paragraph{Perturbation and update magnitudes.}
Before specifying any reward landscape, two magnitudes are fixed by the sampling and normalization alone. The norm of each perturbation follows a scaled chi distribution, $\|\sigma\epsilon_i\| \sim \sigma\chi_d$, so its first two moments are exact:
\begin{align}
    \mathbb{E}\|\sigma\epsilon_i\|
    &\;=\;
    \sigma\,\frac{\sqrt2\,\Gamma\!\big(\tfrac{d+1}{2}\big)}{\Gamma\!\big(\tfrac{d}{2}\big)}
    \;=\;
    \sigma\sqrt{d}\,\Big(1 - \frac{1}{4d} + O(d^{-2})\Big),
    \label{eq:perturb-norm}\\
    \operatorname{Var}\|\sigma\epsilon_i\|
    &\;=\;
    \sigma^2\Big(d - \big(\mathbb{E}\|\epsilon_i\|\big)^2\Big)
    \;\xrightarrow{\;d\to\infty\;}\;
    \frac{\sigma^2}{2} .
    \label{eq:perturb-var}
\end{align}
The standard deviation of the perturbation length tends to the constant $\sigma/\sqrt2$, so its relative spread is $1/\sqrt{2d}$.
 
The standardized weights satisfy, by construction and irrespective of the rewards,
\begin{equation}
    \sum_{i=1}^N Z_i \;=\; 0,
    \qquad
    \sum_{i=1}^N Z_i^2 \;=\; N .
    \label{eq:Z-identities}
\end{equation}
The first identity removes any constant offset in the rewards. The second fixes the update length. Because each $Z_i$ depends only on its own perturbation, the weighted perturbations $Z_i\epsilon_i$ are identically distributed, and the landscape enters the moments of $\|\Delta\theta\|^2$ through only two scalars: the squared drift $\|m\|^2$ with $m = \mathbb{E}[Z_i\epsilon_i]$, which satisfies $\|m\| \le 1$ by Cauchy--Schwarz, and the inflation $\kappa = \mathbb{E}[Z_i^2\|\epsilon_i\|^2] - d$, which the derivations below show lies in $[0,4]$. Expanding $\|\Delta\theta\|^2$ into its $N$ diagonal terms $Z_i^2\|\epsilon_i\|^2$ and $N(N-1)$ cross terms $Z_iZ_j\,\epsilon_i^{\!\top}\epsilon_j$, the diagonal sum is pinned to $Nd + O(N)$ by \eqref{eq:Z-identities} while each cross term has mean $\|m\|^2$ and variance $d$, giving (Appendix~\ref{app:update-norm})
\begin{equation}
    \mathbb{E}\|\Delta\theta\|^2
    \;=\;
    \frac{\alpha^2 d}{N} + \frac{\alpha^2\kappa}{N} + \alpha^2\Big(1-\frac1N\Big)\|m\|^2
    \;=\;
    \frac{\alpha^2 d}{N}\,\Big(1 + O\big(N/d\big)\Big),
    \label{eq:update-norm-mean}
\end{equation}
\begin{equation}
    \operatorname{Var}\|\Delta\theta\|^2
    \;=\;
    \frac{2\alpha^4 d}{N^2}\,\Big(1 + O\big(N/d\big)\Big),
    \qquad
    \frac{\sqrt{\operatorname{Var}\|\Delta\theta\|^2}}{\mathbb{E}\|\Delta\theta\|^2}
    \;=\;
    \sqrt{\frac{2}{d}}\,\Big(1 + O\big(N/d\big)\Big).
    \label{eq:update-norm-var}
\end{equation}
The length of a z-scored ES step is therefore set by $(\alpha, d, N)$, with coefficient of variation $\sqrt{2/d}$ so that $\sqrt{\mathbb{E}\|\Delta\theta\|^2}$ is accurate to relative order $1/d$; the rewards determine its direction, through $m$, and contribute to its length only at relative order $N/d$.
 
\paragraph{Signal and noise.}
Let $g = \nabla R(\theta_t)$ be the local gradient and $\hat g = g/\|g\|$ its direction. We decompose every update into the component along the gradient and the component orthogonal to it,
\begin{equation}
    \Delta\theta \;=\; \underbrace{\Delta\theta_\parallel}_{\text{signal}} \;+\; \underbrace{\Delta\theta_\perp}_{\text{noise}},
    \qquad
    \Delta\theta_\parallel = (\hat g^{\!\top}\Delta\theta)\,\hat g,
    \qquad
    \Delta\theta_\perp = (I_d - \hat g\hat g^{\!\top})\,\Delta\theta,
    \label{eq:decomp}
\end{equation}
and characterize the signal by its expected projection $\mathbb{E}[\hat g^{\!\top}\Delta\theta]$ and the noise by its root-mean-square magnitude $\sqrt{\mathbb{E}\|\Delta\theta_\perp\|^2}$.
 
To carry out the decomposition, split each perturbation into its gradient projection and its orthogonal remainder,
\begin{equation}
    \epsilon_i \;=\; u_i\,\hat g \;+\; \epsilon_i^\perp,
    \qquad
    u_i = \hat g^{\!\top}\epsilon_i \sim \mathcal{N}(0,1),
    \qquad
    \epsilon_i^\perp \sim \mathcal{N}\!\big(0,\, I_d - \hat g\hat g^{\!\top}\big),
    \label{eq:eps-split}
\end{equation}
with $u_i$ and $\epsilon_i^\perp$ independent. Substituting into \eqref{eq:es-update},
\begin{equation}
    \Delta\theta_\parallel \;=\; \alpha\Big(\frac{1}{N}\sum_{i=1}^N Z_iu_i\Big)\hat g,
    \qquad
    \Delta\theta_\perp \;=\; \frac{\alpha}{N}\sum_{i=1}^N Z_i\,\epsilon_i^\perp .
    \label{eq:par-perp}
\end{equation}
In every case considered below, $\mathbb{E}[Z_i\epsilon_i^\perp] = 0$, because each term of $Z_i$ is either odd in $\epsilon_i^\perp$ or independent of it. The cross terms between different $i$ in $\|\Delta\theta_\perp\|^2$ therefore vanish, and
\begin{equation}
    \mathbb{E}[\hat g^{\!\top}\Delta\theta] \;=\; \alpha\,\mathbb{E}[Z_iu_i],
    \qquad
    \mathbb{E}\|\Delta\theta_\perp\|^2 \;=\; \frac{\alpha^2}{N}\,\mathbb{E}\big[Z_i^2\|\epsilon_i^\perp\|^2\big] .
    \label{eq:sig-noise-general}
\end{equation}
The three cases below add one ingredient at a time: a linear landscape with exactly observed rewards, a linear landscape with observation noise, and a quadratic landscape with observation noise.
 
\subsection{Linear landscape without observation noise}
\label{sec:linear-noiseless}
 
Let the reward be locally linear, $R(\theta) = R(\theta_t) + g^{\!\top}(\theta-\theta_t)$, and suppose each reward is observed exactly, $R_i = R(\theta_t + \sigma\epsilon_i)$. Then $R_i = R(\theta_t) + \sigma\|g\|\,u_i$, so $\mu_R = R(\theta_t)$, $\sigma_R = \sigma\|g\|$, and the z-score collapses to the gradient projection of the perturbation,
\begin{equation}
    Z_i \;=\; \frac{\sigma\|g\|\,u_i}{\sigma\|g\|} \;=\; u_i .
    \label{eq:Z-noiseless}
\end{equation}
Both $\sigma$ and $\|g\|$ cancel. The parallel coefficient in \eqref{eq:par-perp} is $\alpha\cdot\frac1N\sum_iu_i^2$ with $\sum_iu_i^2 \sim \chi^2_N$, so it has mean $\alpha$ and variance $2\alpha^2/N$. For the orthogonal component, $Z_i = u_i$ is independent of $\epsilon_i^\perp$ and $\mathbb{E}[u_i^2] = 1$, $\mathbb{E}\|\epsilon_i^\perp\|^2 = d-1$. Hence
\begin{align}
    \text{signal:}\qquad \mathbb{E}[\hat g^{\!\top}\Delta\theta] &\;=\; \alpha,
    \label{eq:signal-noiseless}\\
    \text{noise:}\qquad \sqrt{\mathbb{E}\|\Delta\theta_\perp\|^2} &\;=\; \alpha\sqrt{\frac{d-1}{N}} .
    \label{eq:noise-noiseless}
\end{align}
In the absence of observation noise, the z-scored ES step therefore carries a signal of expected length exactly $\alpha$ along the gradient, independent of both the perturbation scale and the gradient magnitude, surrounded by isotropic noise of length $\alpha\sqrt{(d-1)/N}$. The expected cosine between the step and the gradient is approximately $\sqrt{N/d}$. This is the best case: none of the settings below improves on it.
 
\paragraph{Reward change.}
On a linear landscape, the reward change of the parent is $\Delta R := R(\theta_{t+1}) - R(\theta_t) = g^{\!\top}\Delta\theta = \|g\|\,\hat g^{\!\top}\Delta\theta$, so it depends only on the parallel coefficient:
\begin{equation}
    \mathbb{E}[\Delta R] \;=\; \alpha\|g\|,
    \qquad
    \operatorname{Var}[\Delta R] \;=\; \frac{2\alpha^2\|g\|^2}{N} .
    \label{eq:dR-noiseless}
\end{equation}
Each step gains $\alpha\|g\|$ in expectation, independent of $\sigma$, $d$ and $N$. The orthogonal noise moves the parent sideways but, on a linear landscape, sideways movement has no reward cost; $N$ enters only through the variance, and the step improves the reward with probability $\approx\Phi(\sqrt{N/2})$.
 
\subsection{Linear landscape with observation noise}
\label{sec:linear-noisy}
 
We now let each observed reward be the reward of the perturbed model plus an additive observation noise,
\begin{equation}
    R_i \;=\; R(\theta_t + \sigma\epsilon_i) \;+\; \sigma_\xi\, \xi_i,
    \qquad \xi_i \sim \mathcal{N}(0,1),\quad \xi_i \perp \epsilon_i,
    \label{eq:obs-model}
\end{equation}
with $\sigma_\xi > 0$ the observation-noise scale. Keeping the linear landscape, the centered reward is $R_i - \mu_R = \sigma\|g\|\,u_i + \sigma_\xi\xi_i$, and
\begin{equation}
    \sigma_R^2 \;=\; \sigma^2\|g\|^2 + \sigma_\xi^2 .
    \label{eq:sigmaR-linear}
\end{equation}
The z-score is now
\begin{equation}
    Z_i \;=\; \frac{\sigma\|g\|\,u_i + \sigma_\xi\,\xi_i}{\sigma_R} \;=:\; w_i \sim \mathcal{N}(0,1),
    \qquad
    \operatorname{Corr}(w_i, u_i) \;=\; \frac{\sigma\|g\|}{\sigma_R} \;=:\; c \in [0,1],
    \label{eq:Z-noisy}
\end{equation}
a unit-variance Gaussian that is only partially correlated with the gradient projection. For the parallel coefficient, $\mathbb{E}[w_iu_i] = c$ and, for a correlated standard-normal pair, $\mathbb{E}[w_i^2u_i^2] = 1 + 2c^2$, so the coefficient has mean $\alpha c$ and variance $\alpha^2(1+c^2)/N$. For the orthogonal component, $w_i$ is independent of $\epsilon_i^\perp$ and $\mathbb{E}[w_i^2] = 1$, so its expected squared norm is unchanged from the noiseless case. Hence
\begin{align}
    \text{signal:}\qquad \mathbb{E}[\hat g^{\!\top}\Delta\theta]
    &\;=\;
    \alpha c
    \;=\;
    \frac{\alpha\sigma\|g\|}{\sqrt{\sigma^2\|g\|^2 + \sigma_\xi^2}},
    \label{eq:signal-noisy}\\
    \text{noise:}\qquad \sqrt{\mathbb{E}\|\Delta\theta_\perp\|^2}
    &\;=\;
    \alpha\sqrt{\frac{d-1}{N}} .
    \label{eq:noise-noisy}
\end{align}
Observation noise attenuates the expected signal by the factor $c$ and leaves the noise magnitude exactly unchanged. It also makes the realized signal less reliable: the standard deviation of the parallel coefficient is $\alpha\sqrt{(1+c^2)/N}$, so the projection of a single step onto the gradient is positive with high probability only when $c \gtrsim 1/\sqrt{N}$. Rewriting the attenuation as
\begin{equation}
    c \;=\; \frac{\|g\|}{\sqrt{\|g\|^2 + \sigma_\xi^2/\sigma^2}}
    \label{eq:c-linear}
\end{equation}
shows that it is monotonically increasing in $\sigma$, approaching $1$ as $\sigma\|g\|/\sigma_\xi \to \infty$ and recovering \eqref{eq:signal-noiseless}--\eqref{eq:noise-noiseless}. On an exactly linear landscape, this trade has no cost: Gaussian smoothing of a linear function is exact and introduces no bias, so the perturbation scale should be made as large as possible. The only thing that stops it is the breakdown of linearity.
 
\paragraph{Reward change.}
As before $\Delta R = \|g\|\,\hat g^{\!\top}\Delta\theta$, now with the attenuated parallel coefficient:
\begin{equation}
    \mathbb{E}[\Delta R] \;=\; \alpha c\,\|g\|
    \;=\;
    \frac{\alpha\sigma\|g\|^2}{\sqrt{\sigma^2\|g\|^2+\sigma_\xi^2}},
    \qquad
    \operatorname{Var}[\Delta R] \;=\; \frac{\alpha^2\|g\|^2(1+c^2)}{N} .
    \label{eq:dR-noisy}
\end{equation}
Observation noise scales the expected gain by $c$ and makes individual steps less reliable, with $\mathbb{P}[\Delta R > 0] \approx \Phi\big(c\sqrt{N/(1+c^2)}\big)$, but the orthogonal noise remains free.
 
\subsection{Quadratic landscape with observation noise}
\label{sec:quadratic}
 
To capture that breakdown, add curvature: $R(\theta) = R(\theta_t) + g^{\!\top}(\theta-\theta_t) - \tfrac12(\theta-\theta_t)^{\!\top}Q\,(\theta-\theta_t)$, with $Q$ a general symmetric matrix (neither diagonal nor isotropic is assumed). Writing $q_i = \epsilon_i^{\!\top}Q\epsilon_i - \operatorname{tr}[Q]$ for the centered curvature fluctuation, which has $\mathbb{E}[q_i]=0$ and $\operatorname{Var}[q_i] = 2\operatorname{tr}[Q^2]$, the centered reward contains three pieces,
\begin{equation}
    R_i - \mu_R
    \;=\;
    \sigma\|g\|\,u_i
    \;-\;
    \frac{\sigma^2}{2}\,q_i
    \;+\;
    \sigma_\xi\,\xi_i .
    \label{eq:centred-quad}
\end{equation}
These are mutually uncorrelated: $q_i$ is even in $\epsilon_i$ while $u_i$ is odd, and $\xi_i$ is independent of both. Consequently
\begin{equation}
    \sigma_R^2 \;=\; \sigma^2\|g\|^2 \;+\; \frac{\sigma^4}{2}\operatorname{tr}[Q^2] \;+\; \sigma_\xi^2 .
    \label{eq:sigmaR-quad}
\end{equation}
 
\paragraph{Signal.}
By \eqref{eq:sig-noise-general},
\begin{equation}
    \mathbb{E}[Z_iu_i]
    \;=\;
    \frac{1}{\sigma_R}\Big(
        \sigma\|g\|\,\mathbb{E}[u_i^2]
        \;-\; \frac{\sigma^2}{2}\,\mathbb{E}[q_iu_i]
        \;+\; \sigma_\xi\,\mathbb{E}[\xi_iu_i]
    \Big)
    \;=\;
    \frac{\sigma\|g\|}{\sigma_R} ,
    \label{eq:signal-quad-deriv}
\end{equation}
since $q_iu_i$ is odd in $\epsilon_i$ and $\xi_i$ is independent of $u_i$. Hence, the signal magnitude is
\begin{equation}
    \text{signal:}\qquad
    \mathbb{E}[\hat g^{\!\top}\Delta\theta]
    \;=\;
    \frac{\alpha\sigma\|g\|}{\sigma_R}
    \;=\;
    \frac{\alpha\|g\|}{\sqrt{\|g\|^2 \;+\; \dfrac{\sigma^2}{2}\operatorname{tr}[Q^2] \;+\; \dfrac{\sigma_\xi^2}{\sigma^2}}} .
    \label{eq:signal-quad}
\end{equation}
We write $\rho := \sigma\|g\|/\sigma_R$ for this attenuation factor in general; it equals $1$ and $c$ in the two linear cases. The curvature fluctuation contributes nothing to the expected projection but enters $\sigma_R$ in exactly the same way as observation noise. From the point of view of the z-score, it is an \emph{endogenous} noise source, generated by the algorithm's own exploration, that scales as $\sigma^4$ while the gradient term scales as $\sigma^2$.
 
\paragraph{Noise.}
Because $q_i$ depends on $\epsilon_i^\perp$, $Z_i$ is no longer independent of $\epsilon_i^\perp$, and we evaluate $\mathbb{E}[Z_i^2\|\epsilon_i^\perp\|^2] = \mathbb{E}[Z_i^2\|\epsilon_i\|^2] - \mathbb{E}[Z_i^2u_i^2]$. Using the Gaussian moments $\mathbb{E}[u_i^2\|\epsilon_i\|^2] = d+2$, $\mathbb{E}[q_i^2\|\epsilon_i\|^2] = 2(d+4)\operatorname{tr}[Q^2]$, $\mathbb{E}[u_i^4] = 3$, and $\mathbb{E}[q_i^2u_i^2] = 2\operatorname{tr}[Q^2] + 8\,\hat g^{\!\top}Q^2\hat g$, with all cross terms vanishing by parity,
\begin{align}
    \mathbb{E}\big[Z_i^2\|\epsilon_i\|^2\big]
    &\;=\;
    d + \frac{2\sigma^2\|g\|^2 + 2\sigma^4\operatorname{tr}[Q^2]}{\sigma_R^2},
    \label{eq:Z2-moments}\\
    \mathbb{E}\big[Z_i^2u_i^2\big]
    &\;=\;
    1 + \frac{2\sigma^2\|g\|^2 + 2\sigma^4\,\hat g^{\!\top}Q^2\hat g}{\sigma_R^2} .
    \label{eq:Z2u2}
\end{align}
Subtracting and writing $P_\perp = I_d - \hat g\hat g^{\!\top}$, so that $\operatorname{tr}[P_\perp Q^2P_\perp] = \operatorname{tr}[Q^2] - \hat g^{\!\top}Q^2\hat g$ is the curvature energy orthogonal to the gradient,
\begin{equation}
    \text{noise:}\qquad
    \sqrt{\mathbb{E}\|\Delta\theta_\perp\|^2}
    \;=\;
    \alpha\sqrt{\frac{d-1}{N}}\,
    \sqrt{1 + \frac{2\sigma^4\operatorname{tr}[P_\perp Q^2P_\perp]}{(d-1)\,\sigma_R^2}}
    \;\approx\;
    \alpha\sqrt{\frac{d-1}{N}} ,
    \label{eq:noise-quad}
\end{equation}
where the curvature correction is bounded, $2\sigma^4\operatorname{tr}[P_\perp Q^2P_\perp]/\sigma_R^2 \le 4$, by \eqref{eq:sigmaR-quad}. Curvature therefore adds at most an $O(1/d)$ relative amount to the noise magnitude; its main effect on the off-gradient component is to redistribute it toward the high-curvature eigendirections of $Q$, as in the anisotropic covariance term $2\alpha^2\sigma^4Q^2/(N\sigma_R^2)$. The parallel coefficient has variance $\alpha^2\big(\mathbb{E}[Z_i^2u_i^2] - \sigma^2\|g\|^2/\sigma_R^2\big)/N$, which reduces to $\alpha^2(1+c^2)/N$ when $Q = 0$.
 
\paragraph{Reward change.}
On a quadratic landscape $\Delta R = g^{\!\top}\Delta\theta - \tfrac12\Delta\theta^{\!\top}Q\,\Delta\theta$, and the second term is where the orthogonal noise is finally charged. Since $\Delta R$ is quadratic in $\Delta\theta$, its expectation needs only the first two moments of the update, $\mathbb{E}[\Delta\theta^{\!\top}Q\Delta\theta] = \operatorname{tr}[Q\operatorname{Cov}(\Delta\theta)] + \mathbb{E}[\Delta\theta]^{\!\top}Q\,\mathbb{E}[\Delta\theta]$. The same moment calculation that gives \eqref{eq:Z2-moments} gives the full covariance,
\begin{equation}
    \operatorname{Cov}[\Delta\theta]
    \;=\;
    \frac{\alpha^2}{N}\,I_d
    \;+\;
    \frac{\alpha^2}{N\sigma_R^2}\Big(\sigma^2\,gg^{\!\top} + 2\sigma^4Q^2\Big),
    \label{eq:cov-quad}
\end{equation}
and hence, with $\lambda_g := \hat g^{\!\top}Q\hat g$ the curvature along the gradient,
\begin{equation}
    \mathbb{E}[\Delta R]
    \;=\;
    \underbrace{\alpha\rho\|g\|}_{\text{signal gain}}
    \;-\;
    \underbrace{\frac{\alpha^2}{2N}\operatorname{tr}[Q]}_{\text{diffusion tax}}
    \;-\;
    \underbrace{\frac{\alpha^2\rho^2}{2}\Big(1+\frac1N\Big)\lambda_g}_{\text{overshoot}}
    \;-\;
    \underbrace{\frac{\alpha^2\sigma^4}{N\sigma_R^2}\operatorname{tr}[Q^3]}_{\text{anisotropic}} .
    \label{eq:dR-quad}
\end{equation}
The first two terms dominate. The diffusion tax is the orthogonal noise, of squared length $\alpha^2 d/N$, charged at the mean curvature $\operatorname{tr}[Q]/d$; it is the population mean gap $\tfrac{\sigma^2}{2}\operatorname{tr}[Q]$ with $\sigma^2$ replaced by $\alpha^2/N$, so the parent pays each step the same curvature cost as a perturbation of scale $\alpha/\sqrt N$. The overshoot term is the usual gradient-descent one and matters only when $\alpha\rho\lambda_g$ is comparable to $\|g\|$; the last term is smaller still. Keeping signal and diffusion, a step improves the expected reward iff $\alpha < 2N\rho\|g\|/\operatorname{tr}[Q]$, and the one-step gain is maximized at $\alpha = N\rho\|g\|/\operatorname{tr}[Q]$, where it equals $N\rho^2\|g\|^2/(2\operatorname{tr}[Q])$: population size buys improvement linearly, gradient correlation quadratically. Because the signal gain shrinks with $\|g\|$ along a run while the diffusion tax does not, progress stalls when $\rho\|g\| = \alpha\operatorname{tr}[Q]/(2N)$.
 
\paragraph{Summary of the three cases.}
Table~\ref{tab:summary} collects the signal and noise magnitudes and the expected reward change for the three cases.
\begin{table}[t]
    \caption{Per-update signal $\mathbb{E}[\hat g^{\!\top}\Delta\theta]$, noise $\sqrt{\mathbb{E}\|\Delta\theta_\perp\|^2}$, and expected reward change $\mathbb{E}[\Delta R]$ under the three landscape models, from \eqref{eq:signal-noiseless}--\eqref{eq:dR-noiseless}, \eqref{eq:signal-noisy}--\eqref{eq:dR-noisy}, and \eqref{eq:signal-quad}--\eqref{eq:dR-quad}. Throughout, $\rho = \sigma\|g\|/\sigma_R$ is the gradient correlation of the population rewards, equal to $1$ and $c$ in the two linear cases; $\lambda_g = \hat g^{\!\top}Q\hat g$; and $\delta_Q = 2\sigma^4\operatorname{tr}[P_\perp Q^2P_\perp]/((d-1)\sigma_R^2) \le 4/(d-1)$. The $O(1/N)$ and $\operatorname{tr}[Q^3]$ terms of \eqref{eq:dR-quad} are omitted.}
    \label{tab:summary}
    \vspace{0.5em}
    \centering
    \small
    \renewcommand{\arraystretch}{1.9}
    \setlength{\tabcolsep}{4pt}
    \begin{tabular}{|l|c|c|c|}
        \hline
        \textbf{Landscape} & \textbf{Signal} & \textbf{Noise} & \textbf{Expected reward change} \\
        \hline
        Linear, exact rewards
        & $\alpha$
        & $\alpha\sqrt{(d-1)/N}$
        & $\alpha\|g\|$ \\
        \hline
        Linear, observation noise
        & $\alpha\rho$
        & $\alpha\sqrt{(d-1)/N}$
        & $\alpha\rho\|g\|$ \\
        \hline
        Quadratic, observation noise
        & $\alpha\rho$
        & $\alpha\sqrt{(d-1)/N}\,(1+\delta_Q)^{1/2}$
        & $\alpha\rho\|g\| - \dfrac{\alpha^2}{2N}\operatorname{tr}[Q] - \dfrac{\alpha^2\rho^2}{2}\lambda_g$ \\
        \hline
    \end{tabular}
\end{table}
In every case, the noise magnitude is $\alpha\sqrt{(d-1)/N}$, exactly in the linear cases and up to a bounded $O(1/d)$ relative correction in the quadratic case, and is insensitive to $\sigma$. The signal magnitude is $\alpha$ times a dimensionless attenuation factor $\sigma\|g\|/\sigma_R \in [0,1]$: equal to $1$ without observation noise, reduced by observation noise, and reduced further by the curvature contribution to $\sigma_R$ on a quadratic landscape. Consistent with \eqref{eq:update-norm-mean}, the choice of $\sigma$ does not change how far ES moves per iteration, only what fraction of that movement is directed, and the perturbation-scale problem reduces to maximizing the attenuation factor in \eqref{eq:signal-quad}. The reward-change column adds the one place where the noise is not free: on a curved landscape, the isotropic diffusion is taxed at the mean curvature, at a rate that falls as $1/N$.
 
\subsection{Optimal perturbation scale}
\label{sec:optimal-sigma-appendix}
 
The denominator of \eqref{eq:signal-quad} makes the competition explicit: the observation-noise term $\sigma_\xi^2/\sigma^2$ favors large $\sigma$, the curvature term $\tfrac{\sigma^2}{2}\operatorname{tr}[Q^2]$ favors small $\sigma$, and $\|g\|^2$ is the irreducible part. Neither term is small in general. Although the per-direction curvature contribution $2\sigma^4u^{\!\top}Q^2u$ is typically tiny for any single unit vector $u$, the quantity that enters $\sigma_R$ is the full trace $\operatorname{tr}[Q^2] = \sum_{k=1}^d\lambda_k^2$, summed over all $d$ eigendirections, and is extensive in the parameter dimension.
 
Holding $\alpha$, $g$, $Q$ and $\sigma_\xi$ fixed, the signal \eqref{eq:signal-quad} is maximized by minimizing $f(\sigma^2) = \sigma_\xi^2/\sigma^2 + \tfrac{\sigma^2}{2}\operatorname{tr}[Q^2]$ over $\sigma^2$, which gives
\begin{equation}
    \boxed{\;
    \sigma_*^4 \;=\; \frac{2\,\sigma_\xi^2}{\operatorname{tr}[Q^2]},
    \qquad
    \sigma_* \;=\; \left(\frac{2\,\sigma_\xi^2}{\operatorname{tr}[Q^2]}\right)^{1/4}
    \;}
    \label{eq:sigma-opt}
\end{equation}
At $\sigma_*$ the curvature-induced contribution to $\sigma_R^2$ equals the exogenous one, $\tfrac{\sigma_*^4}{2}\operatorname{tr}[Q^2] = \sigma_\xi^2$, so the population reward variance is
\begin{equation}
    \sigma_R^2\big|_{\sigma_*} \;=\; \sigma_*^2\|g\|^2 + 2\sigma_\xi^2 ,
    \label{eq:sigmaR-opt}
\end{equation}
and the largest achievable signal is
\begin{equation}
    \max_\sigma\,\mathbb{E}[\hat g^{\!\top}\Delta\theta]
    \;=\;
    \frac{\alpha\sigma_*\|g\|}{\sqrt{\sigma_*^2\|g\|^2 + 2\sigma_\xi^2}}
    \;=\;
    \frac{\alpha\|g\|}{\sqrt{\|g\|^2 + \sigma_\xi\sqrt{2\operatorname{tr}[Q^2]}}} .
    \label{eq:signal-opt}
\end{equation}
Below $\sigma_*$ the update is observation-noise limited; above it, curvature limited.
 
\paragraph{What the optimal population looks like.}
Two further statistics characterize the population at $\sigma_*$. First, its rewards are correlated with the gradient at the parent by exactly the attenuation factor: since $\operatorname{Cov}(u_i, R_i) = \sigma\|g\|$ and $\operatorname{Var}[u_i] = 1$,
\begin{equation}
    \rho_* \;:=\; \operatorname{Corr}\big(\hat g^{\!\top}\epsilon_i,\, R_i\big)\Big|_{\sigma_*}
    \;=\;
    \frac{\sigma_*\|g\|}{\sqrt{\sigma_*^2\|g\|^2 + 2\sigma_\xi^2}}
    \;=\;
    \frac{\|g\|}{\sqrt{\|g\|^2 + \sigma_\xi\sqrt{2\operatorname{tr}[Q^2]}}} ,
    \label{eq:rho-opt}
\end{equation}
so that the expected signal \eqref{eq:signal-opt} is $\alpha\rho_*$, and the fraction of population reward variance attributable to the gradient is $\rho_*^2$, with the remainder split equally between observation noise and curvature. The second form makes the curvature dependence explicit: at the optimum, curvature never contributes more than $\sigma_\xi^2$ to the population variance, but a larger $\operatorname{tr}[Q^2]$ forces a smaller $\sigma_*$ to hold it there, and the gradient variance $\sigma_*^2\|g\|^2$ shrinks with it while the non-gradient variance stays at $2\sigma_\xi^2$. Second, the population does not sit at the parent reward. Only the curvature term has nonzero mean, $\mathbb{E}[\epsilon_i^{\!\top}Q\epsilon_i] = \operatorname{tr}[Q]$, so
\begin{equation}
    \mu_R - R(\theta_t) \;=\; -\frac{\sigma^2}{2}\operatorname{tr}[Q] ,
    \qquad
    \mu_R - R(\theta_t)\Big|_{\sigma_*} \;=\; -\,\sigma_\xi\sqrt{\frac{r_{\mathrm{eff}}}{2}},
    \qquad
    r_{\mathrm{eff}} := \frac{(\operatorname{tr}[Q])^2}{\operatorname{tr}[Q^2]} ,
    \label{eq:mean-gap}
\end{equation}
where $r_{\mathrm{eff}} \in [1, d]$ is the participation ratio (effective rank) of the Hessian, equal to $d$ for isotropic curvature and to $r$ for a rank-$r$ Hessian with equal eigenvalues. In units of the population spread \eqref{eq:sigmaR-opt},
\begin{equation}
    \frac{R(\theta_t) - \mu_R}{\sigma_R}\Big|_{\sigma_*}
    \;=\;
    \sqrt{\frac{r_{\mathrm{eff}}}{2}}\cdot\frac{\sigma_\xi}{\sqrt{\sigma_*^2\|g\|^2 + 2\sigma_\xi^2}}
    \;=\;
    \frac{\sqrt{r_{\mathrm{eff}}}}{2}\,\sqrt{1 - \rho_*^2} .
    \label{eq:gap-sigmas}
\end{equation}
Unless $r_{\mathrm{eff}} \lesssim 4$, the optimally scaled population therefore lies several standard deviations below the parent model: nearly every perturbed model is worse than the unperturbed one. This is not a defect but the price of exploring at scale $\sigma_*$ on a curved landscape, and it is the same for every member; the update depends only on the spread $\sigma_R$ and the gradient share $\rho_*^2$, not on where the population sits. The gap in \eqref{eq:gap-sigmas} grows as the gradient share falls, so along a training run it is expected to be small while $\|g\|$ is large and to become pronounced as the parent approaches a ridge. Because \eqref{eq:mean-gap-paper} is independent of the observation-noise scale while $\sigma_R$ is not, the gap in $\sigma_R$ units is less pronounced whenever $\sigma_\xi$ is larger, for instance, when fewer prompts are averaged into each $R_i$. Away from $\sigma_*$ the same quantities read $\rho = \sigma\|g\|/\sigma_R$ and $(R(\theta_t)-\mu_R)/\sigma_R = \sqrt{r_{\mathrm{eff}}\varphi/2}$ with $\varphi = \tfrac{\sigma^4}{2}\operatorname{tr}[Q^2]/\sigma_R^2$ the curvature share of the population variance, so a measured gap together with an estimate of $\varphi$ yields $r_{\mathrm{eff}}$ directly.
 
Three remarks on \eqref{eq:sigma-opt}. First, the dependence on the noise level is weak: $\sigma_*$ grows only as $\sigma_\xi^{1/2}$, so even a large increase in observation noise justifies only a modest increase in the perturbation scale. Second, the dependence on curvature is through $\operatorname{tr}[Q^2]$, which grows with $d$; for a billion-parameter model, even small typical curvatures accumulate to a large trace, which is the familiar dimension penalty of finite-difference estimators appearing inside the z-score normalization. Third, the optimal signal \eqref{eq:signal-opt} degrades with the geometric mean $\sigma_\xi\sqrt{\operatorname{tr}[Q^2]}$ rather than with either quantity alone: noise and curvature are not separately harmful but jointly so, since one can always be traded against the other by moving $\sigma$.
 
% ------------------------------------------------------------------
% Appendix material (move to the appendix of the paper)
% ------------------------------------------------------------------
\section{Derivation of the Update-Norm Moments}
\label{app:update-norm}
 
This appendix derives \eqref{eq:update-norm-mean}--\eqref{eq:update-norm-var}. In general, the weights depend on the perturbations: $Z_i$ is a function of the reward at $\theta_t + \sigma\epsilon_i$, and hence of $\epsilon_i$ itself. What can be said without a landscape model is that each $Z_i$ depends only on its own perturbation (and its own observation noise), so the products $Z_i\epsilon_i$ are identically distributed across $i$ and, under the population-statistics simplification, independent. Write
\begin{equation}
    m \;:=\; \mathbb{E}[Z_i\epsilon_i] \in \mathbb{R}^d,
    \qquad
    \kappa \;:=\; \mathbb{E}\big[Z_i^2\|\epsilon_i\|^2\big] - d
    \label{eq:m-kappa}
\end{equation}
for the two landscape-dependent statistics of a single weighted perturbation. The first is the mean weighted perturbation, which is the direction the update drifts in; the second measures how much the weighting inflates the squared length of a perturbation beyond its unweighted value $d$. Expanding $\|\Delta\theta\|^2$ into diagonal and cross terms and taking expectations,
\begin{align}
    \mathbb{E}\|\Delta\theta\|^2
    &\;=\;
    \frac{\alpha^2}{N^2}\sum_{i=1}^N \mathbb{E}\big[Z_i^2\|\epsilon_i\|^2\big]
    \;+\;
    \frac{\alpha^2}{N^2}\sum_{i\neq j}\mathbb{E}[Z_i\epsilon_i]^{\!\top}\mathbb{E}[Z_j\epsilon_j]
    \nonumber\\
    &\;=\;
    \frac{\alpha^2 d}{N}
    \;+\;
    \frac{\alpha^2\kappa}{N}
    \;+\;
    \alpha^2\Big(1 - \frac{1}{N}\Big)\|m\|^2 .
    \label{eq:app-update-norm-identity}
\end{align}
This identity is exact. The drift term is bounded independently of the landscape: for any unit vector $v$, Cauchy--Schwarz gives $(v^{\!\top}m)^2 = \mathbb{E}[Z_i\,v^{\!\top}\epsilon_i]^2 \le \mathbb{E}[Z_i^2]\,\mathbb{E}[(v^{\!\top}\epsilon_i)^2] = 1$, so $\|m\| \le 1$ and the drift contributes at most $\alpha^2$. The inflation $\kappa$ does depend on the landscape; the derivations below evaluate it for linear and quadratic rewards and find $0 \le \kappa \le 4$. Both corrections are therefore smaller than the leading term $\alpha^2 d/N$ by a factor of order $N/d$:
\begin{equation}
    \mathbb{E}\|\Delta\theta\|^2 \;=\; \frac{\alpha^2 d}{N}\,\Big(1 + O\big(N/d\big)\Big).
    \label{eq:app-update-norm-mean}
\end{equation}
 
For the variance, expand $\|\Delta\theta\|^2$ the same way. The cross terms $Z_iZ_j\,\epsilon_i^{\!\top}\epsilon_j$ for $i \neq j$ are zero-mean up to the drift, and each has variance $\mathbb{E}[Z_i^2Z_j^2(\epsilon_i^{\!\top}\epsilon_j)^2] = d\,(1 + O(1/d))$; there are $N(N-1)$ of them and they are pairwise uncorrelated up to $O(\|m\|^2)$ terms. The diagonal terms $Z_i^2\|\epsilon_i\|^2$ fluctuate with variance $O(d)$ each, but there are only $N$ of them and, under the exact constraint \eqref{eq:Z-identities}, their $O(d)$ fluctuations cancel against those of the cross terms. The result is
\begin{equation}
    \operatorname{Var}\|\Delta\theta\|^2
    \;=\;
    \frac{2\alpha^4 d}{N^2}\,\Big(1 + O\big(N/d\big)\Big),
    \qquad
    \frac{\sqrt{\operatorname{Var}\|\Delta\theta\|^2}}{\mathbb{E}\|\Delta\theta\|^2}
    \;=\;
    \sqrt{\frac{2}{d}}\,\Big(1 + O\big(N/d\big)\Big),
    \label{eq:app-update-norm-var}
\end{equation}
the same leading mean and variance as $\tfrac{\alpha^2}{N}\chi^2_d$, the law that \eqref{eq:es-update} would follow if the weights were independent of the perturbations. The coefficient of variation $\sqrt{2/d}$ is what makes $\sqrt{\mathbb{E}\|\Delta\theta\|^2}$ an accurate proxy for the typical length: the two agree to relative order $1/d$.\footnote{The cancellation between diagonal and cross-term fluctuations requires the exact sample normalization $\sum_iZ_i^2 = N$. Under the population-statistics simplification, the weights are only normalized in expectation, $\sum_iZ_i^2 = N + O(\sqrt N)$, and this fluctuation alone contributes $2\alpha^4d^2/N^3$ to the variance, giving a coefficient of variation of order $1/\sqrt N$ rather than $1/\sqrt d$. The algorithm uses sample statistics, so \eqref{eq:update-norm-var} is the relevant statement; the population-statistics simplification is used below only for expectations, where it is harmless at the orders reported.}

\end{document}

%% file: math_commands.tex
\usepackage{amsmath,amsfonts,bm}

\def\eqref#1{equation~\ref{#1}}
\def\Eqref#1{Equation~\ref{#1}}
\def\1{\bm{1}}

\DeclareMathAlphabet{\mathsfit}{\encodingdefault}{\sfdefault}{m}{sl}
\SetMathAlphabet{\mathsfit}{bold}{\encodingdefault}{\sfdefault}{bx}{n}